\documentclass[journal]{IEEEtran}

\ifCLASSINFOpdf
  \usepackage[pdftex]{graphicx}
  \DeclareGraphicsExtensions{.pdf,.jpeg,.png}
\else
  \usepackage[dvips]{graphicx}
  \DeclareGraphicsExtensions{.eps}
\fi

\usepackage{amsmath,amsfonts,amssymb}
\usepackage{cite}
\usepackage{color}
\usepackage{pstricks,pst-plot}
\usepackage{pst-sigsys}
\usepackage{psfrag}
\usepackage{ifthen}
\usepackage{balance}
\usepackage{multirow}
\usepackage{pbox}
\usepackage[caption=false,font=footnotesize]{subfig}
\usepackage[makeroom]{cancel}
\usepackage{changepage}

\usepackage{mathrsfs}
\usepackage{amsbsy}

\newcommand{\vu}[2]{\mbox{$#1\,\text{#2}$}}

\definecolor{ForestGreen}{rgb}{0.2, 0.6, 0.1}

\newcommand{\psfragoffset}[6]{\psfrag{#1}{#2 \textcolor{#3}{\raisebox{#5em}{\hspace{#4em}#6}}}}
\definecolor{darkgrey}{rgb}{0.3, 0.3, 0.3}

\newcommand{\be}{\begin{equation}}
\newcommand{\ee}{\end{equation}}
\newcommand{\ist}{\hspace*{.3mm}}
\newcommand{\rmv}{\hspace*{-.3mm}}

\newcommand{\bd}[1]{\mathbf{#1}}
\newcommand{\cl}[1]{\mathcal{#1}}
\newcommand{\nn}{\nonumber}

\newcommand{\E}[2]{\mathbb{E}_{#1}\left\{ #2 \right\}}   %expectation
\newcommand{\tr}{\mathrm{tr}}                       %trace of matrix

\newcommand{\conv}{\ast}                            %convolution

\newcommand{\var}[1]{\mathrm{var}\left\{ #1 \right\}}

\newcommand{\agentidxsym}{m}               %the symbol for the agent index
\newcommand{\agentidx}{^{(\agentidxsym)}}  %symbol as it should be used in equations
\newcommand{\mpcidxsym}{k}          %the symbol for the MPC index
\newcommand{\mpcidx}{_{\mpcidxsym}} %symbol as it should be used in equations
\newcommand{\timestepsym}{n}            %the symbol for the discrete time step
\newcommand{\timestep}{_{\timestepsym}} %symbol as it should be used in equations
\newcommand{\trans}{^{\mathrm{T}}}

\newcommand{\FIM}{\pmb{\mathcal{I}}}

\newcommand{\bp}{{\bf p}}

\newcommand{\bJ}{{\bf J}}

\newcommand{\Tp}{{T_\mathrm{p}}}

\newcommand{\SINR}{\mathrm{SINR}}

\newcommand{\pind}[1]{\mathbf{p}_{\mathrm{#1}}}   %a position with an index

\newcommand{\VA}[1]{\mathcal{A}_{#1}}     %Set of VAs (general)
\newcommand{\VAassoc}[1]{\mathcal{A}_{#1, \mathrm{ass}}}     %Set of VAs (associated)
\newcommand{\vm}[1]{\mathbf{#1}}% vector or matrix
\newcommand{\va}[1]{\vm{p}_{#1}}          %a specific VA
\newcommand{\Z}[1]{\mathcal{Z}_{#1}}               %set of measurements
\newcommand{\D}[1]{\mathcal{D}_{#1}}               %set of (true, expected) distances
\newcommand{\fc}{f_\text{c}}  %center frequency
\newcommand{\DT}{\Delta T}

\graphicspath{{./figures/}}

\definecolor{Pblue}{rgb}{0, 0.5, 1}
\definecolor{Fgreen}{rgb}{0.2, 0.6, 0.1}
\definecolor{darkgrey}{rgb}{0.3, 0.3, 0.3}
\definecolor{darkred}{rgb}{0.8, 0, 0}

\begin{document}
\title{Cognitive Indoor Positioning and Tracking using Multipath Channel Information}
\author{Erik Leitinger,~\IEEEmembership{Member,~IEEE}, Simon Haykin,~\IEEEmembership{Life Fellow,~IEEE}, and Klaus Witrisal,~\IEEEmembership{Member,~IEEE} 
	\thanks{E. Leitinger, P. Meissner, and K. Witrisal are with Graz University of Technology, Graz, Austria, email: \{erik.leitinger, paul.meissner, witrisal\}@tugraz.at}}

\maketitle

%%%%%%%%%%%%%%%%%%%%%%%%%%%%%%%%%%%%%%%%%%%%%%%%%%%%%%%%%%%%%%%%%%%%%%%%%%%%%%%%%%%%%%%%%%%%%%%%%%%%%
%%%%%%%%%%%%%%%%%%%%%%%%%%%%%%%%%%%%%%%%%%%%%%%%%%%%%%%%%%%%%%%%%%%%%%%%%%%%%%%%%%%%%%%%%%%%%%%%%%%%%
\begin{abstract}
  This paper presents a \emph{robust} and \emph{accurate} positioning system that adapts its behavior to the surrounding environment, mimicking the capability of the visual brain to filtering out clutter and focusing attention on activity and relevant information. Especially in indoor environments, which are characterized by harsh multipath propagation, robust positioning is still hard to achieve under the constraint of reasonable infrastructural needs. In such environments it is essential to separate relevant from irrelevant information and attain an appropriate uncertainty model for measurements that are used for positioning.
\end{abstract}

%%%%%%%%%%%%%%%%%%%%%%%%%%%%%%%%%%%%%%%%%%%%%%%%%%%%%%%%%%%%%%%%%%%%%%%%%%%%%%%%%%%%%%%%%%%%%%%%%%%%%
%%%%%%%%%%%%%%%%%%%%%%%%%%%%%%%%%%%%%%%%%%%%%%%%%%%%%%%%%%%%%%%%%%%%%%%%%%%%%%%%%%%%%%%%%%%%%%%%%%%%%

%%%%%%%%%%%%%%%%%%%%%%%%%%%%%%%%%%%%%%%%%%%%%%%%%%%%%%%%%%%%%%%%%%%%%%%%%%%%%%%%%%%%%%%%%%%%%%%%%%%%%
%%%%%%%%%%%%%%%%%%%%%%%%%%%%%%%%%%%%%%%%%%%%%%%%%%%%%%%%%%%%%%%%%%%%%%%%%%%%%%%%%%%%%%%%%%%%%%%%%%%%%
\section{Introduction}
\label{sec:Introduction}
\subsection{Motivation and State of the Art}

%General
For radiobased positioning in indoor environments, which are characterized by harsh multipath propagation, it is still elusive to achieve the needed level of accuracy \emph{robustly}\footnote{We define robustness as the percentage of cases in which a system can achieve its given potential accuracy.} under the constraint of reasonable infrastructural needs. In such environments it is essential to separate relevant from irrelevant information and attain an appropriate uncertainty model for measurements that are being used for positioning.

To approach this objective more closely the four basic principles for \emph{human cognition}, namely the \emph{perception-action-cycle (PAC)}, \emph{memory}, \emph{attention} and \emph{intelligence} \cite{Fuster2009} are implemented into the positioning systems as schematically illustrated in Fig.~\ref{fig:CL_Blockdiag}. To encounter all these principles, the concepts of multipath-assisted indoor navigation and tracking (MINT) \cite{MeissnerPhD2014,LeitingerJSAC2015,LeitingerPhD2016,LeitingerGNSS2016} are intertwined with the principles of cognitive dynamic systems (CDS) that were developed in \cite{HaykinPROC2012a, HaykinPROC2012b, Fatemi2014,AmiriNC2014,HaykinPROC2014}. Evidently, a perceptive system has to reason with measurements under \emph{uncertainty} \cite{Pearl1988}, i.e. it has to treat the gained information \emph{probabilistically} \cite{Gregory2005_BLD, SiviaSkilling2006}, but it also has to deliberately take actions on the environment and consequently influence measurements to reason in favor of relevant information instead of irrelevant one. Hence, cognitive processing of measurement data for positioning seems to be a natural choice to overcome such severe impairments. 

MINT exploit specular multipath components (MPCs) that can be associated to the local geometry as illustrated in Fig.~\ref{fig:prob_VAgeometry}. MPCs can be interpreted as signals originiating from additional virtual sources, so-called virtual anchors (VA). These VAs are mirror-images of a physical anchor w.r.t. the flat surfaces as illustrated in Fig.~\ref{fig:prob_VAgeometry} \cite{MeissnerPhD2014,Borish1984,KunischICUWB2003}. This additional \emph{position-related information} can be  utilized from the radio signals. For a proper consideration of uncertainties in the floor plan and to account for the stochastic nature of the radio signals a geometry-based probabilistic environment model (GPEM) and a geometry-based stochastic channel model (GSCM) where introduced in \cite{LeitingerICC2015,LeitingerICC2017,LeitingerTWC2019,LeitingerICC2019}, extending MINT to a simultaneous localization and mapping (SLAM) approach. Such a systems acquires and adapts online information about its surrounding environment and is able to continuously building up a consistent memory in a Bayesian sense. 

The idea of combing MINT with a CDS is to gain control over the observed environment information to (i) provide as much position-related information to the Bayesian state estimator as possible for achieving the highest level of reliability/robustness in position estimation, (ii) to improve the separation between relevant and irrelevant information, and (iii) building up a consistent environment and action memory. By actively planning next control actions on the environment using the Bayesian memory---in sense of waveform adaptation \cite{Kershaw1994, Haykin2011, HaykinPROC2012a, BellJSTSP2015} or mobile agent motor-control \cite{HoffmannTAC2010, JulianJRR2012}---the relevant information-return contained in the signals can be maximized. The information-flow coupling between the perceptor-actor system and the surrounding environment is given by the PAC that plays the key-role when it is coming to gather relevant environment information \cite{Fuster2009,HaykinPROC2014}. 

The core feedback loop of the cognitive dynamic system, the perception-action-cycle resembles the idea of optimally choosing future measurements based on a physical model under reasoning with uncertainty. The principle has been explored by the physics community under the term Bayesian experimental design \cite{Chaloner1995}. This decision-theoretic process gives a mathematical justification for selecting the appropriate optimality criterion under uncertainty that maximizes the utility function of the posterior probability density function, such that new model information of the acquired measurements can be predicted. Information theoretic measures such as the conditional entropy \cite{Cover2006}, the mutual information \cite{Cover2006} or the determinate of the Fisher information matrix \cite{VanTrees1968,Kay1993} are suitable utility functions for this process. The active selection of measurement parameters has a lot in common with cognitive perception and control at the lowest layer. However, it lacks an explicit description of a layered memory structure that, in combination with algorithmic attention leads to an ``intelligent'' behavior of the overall system.

\section{MINT Concepts}
\label{sec:signalmodel}
In this section we review basic elements of MINT \cite{MeissnerWCL2014, LeitingerJSAC2015} starting with the signal model, then discussing the estimation of the MPC parameters, and finally introducing position related information that is of main importance for a proper weighting of the MPC-VA relations in the Bayesian tracking filter. All not-geometrically-modeled propagation effects in the signals, so-called diffuse multipath (DM) \cite{MichelusiTSP2012part1}, constitute interference to the useful position-related information.
 
\subsection{Geometry-based Stochastic Signal Model (GSCM)}\label{sec:channeles}

Our signal model is the following. During time step $n$, a baseband radio signal $s(t) $ is transmitted from the $j$-th physical anchor located at position $\vm{a}_1^{(j)} \in \mathbb{R}^2$, $j  \in \{ 1,\dots,J \} = \mathcal{J}$, to a mobile agent at position $\vm{p}_n \in \mathbb{R}^2$. The corresponding received signal is given as \cite{LeitingerJSAC2015}
\begin{align}\label{eq:rx_signal}
  r_n^{(j)}(t) = \sum_{k=1}^{K_n^{(j)}}\alpha_{k,n}^{(j)} s\big(t-\tau_{k,n}^{(j)}\big) + s(t)\conv\nu_{n}^{(j)}(t) + w(t).
\end{align}
Here, the first term describes the contributions from $K_n^{(j)}$ specular MPCs with complex amplitudes $\alpha_{k,n}^{(j)}$ and delays $\tau_{k,n}^{(j)}$, where $k \in \big\{ 1,\dots,K_n^{(j)}\big\} = \mathcal{K}_n^{(j)}$. The delays $\tau_{k,n}^{(j)}$ correspond to the distances between the agent and the $j$-th physical anchor (for $k = 1$) or the VAs of the $j$-th physical anchor (for $k \in \big\{ 2,\dots,K_n^{(j)}\big\}$). Thus, $\tau_{k,n}^{(j)} = \big\|\vm{p}_n - \vm{a}\mpcidx^{(j)}\big\|\big/c$, where $\vm{a}\mpcidx^{(j)} \in \mathbb{R}^2$ is the position of the respective (physical or virtual) anchor and $c$ is the speed of light. The energy of the transmitted signal $s(t)$ is assumed to be normalized to one. The second term in \eqref{eq:rx_signal} denotes the convolution of $s(t)$ with the diffuse multipath (DM) $\nu_n^{(j)}(t)$, which is modeled as a non-stationary zero-mean Gaussian random process. Considering uncorrelated scattering along the delay axis $\tau$, the auto-correlation function of $\nu_n^{(j)}(t)$ is given by $\mathbb{E}_\nu\big\{ \nu_n^{(j)}(\tau) \nu_n^{(j)\ist*}(u)\big\}  = S_{\nu,n}^{(j)}(\tau) \delta(\tau-u)$, where $S_{\nu,n}^{(j)}(\tau)$ represents the power delay profile of DM. The DM process $\nu_n^{(j)}(t)$ is assumed to be quasi-stationary in the spatial domain, which means that $S_{\nu,n}^{(j)}(\tau)$ does not change in the vicinity of $\vm{p}_n$ \cite{MolischTPROC2009}. Note that the DM component interferes with the useful position-related information. The last term in \eqref{eq:rx_signal}, $w(t)$, is additive white Gaussian noise with double-sided power spectral density $N_0/2$.

\subsection{MPC Parameter Estimation}\label{sec:channelest}

The delays of the MPCs at agent position $\vm{p}_n$ are estimated from the received signals using a sparse Bayesian channel estimator \cite{GrebienLeitingerTSP2021}{}. The algorithm estimates up to a predefined maximum number $M$ of MPCs yielding estimated delays $\hat{\tau}_{m,n}^{(j)}$ and according complex amplitudes $\hat{\alpha}^{(j)}_{m,n}$, with $m \in \{ 1,\dots,M_n^{(j)} \}=\mathcal{M}_n^{(j)}$. The estimated delays are scaled by the speed of light $c$ and used as noisy distance measurements $z_{m,n}^{(j)} = c\hat{\tau}_{m,n}^{(j)}$ in the proposed multipath-assisted SLAM algorithm. Furthermore, in a real-world MINT system, the amplitude estimates $\hat{\alpha}^{(j)}_{m,n}$ (after being associated with the $k$-th anchor) are fed into a higher-level, non-Bayesian algorithm that determines the signal-to-interference-plus-noise power ratio (SINR) between the useful specular MPC and the DM plus noise. This SINR is related to the range standard deviation $\sigma_{m,n}^{(j)}$ (see \cite{MeissnerWCL2014, WitrisalSPM2016} for details). Note that an extension to additional parameters besides the delay (and the corresponding amplitude), as for example the angle-of-arrival and angle-of-departure of the MPCs, is straightforward.

\subsection{Position and Range Uncertainty}\label{sec:SINRest}

As a performance measure and lower bound on the position error we use the Cramer-Rao-Lower Bound (CRLB) defined by the inequality $\mathbb{E}\{||\bp - \hat{\bp}||\} \geq \tr\{ \bJ_{\bp}^{-1}\}$, where $\bJ_{\bp}$ is the equivalent Fisher information matrix (EFIM) \cite{ShenTIT2010part1,ShenTIT2010part2,LeitingerJSAC2015} for the position vector and $\tr\{\cdot\}$ is the trace operator. Assuming no path overlap between MPCs, the EFIM $\bJ_{\bp}$ is formulated for a set of anchors in a canonical form by \cite{LeitingerJSAC2015}
\begin{align}
	\FIM_{\vm{p},n} = \frac{8\pi^2\beta^2}{c^2} \sum_{j=1}^{J}\sum_{k=1}^{K_n^{(j)}} \SINR_{k,n}^{(j)} \FIM_{\mathrm{r}}\left(\phi_{k,n}^{(j)} \right),
	\label{eq:EFIM_pos}
\end{align} 
where $\beta$ denotes the effective (root mean square) bandwidth of $s(t)$ and $\FIM_{\mathrm{r}}(\phi_{k,n}^{(j)})$ is the ranging direction matrix, which is a rank-one matrix with an eigenvector in direction $\phi_{k,n}^{(j)}$ from the agent to the $k$-th VA. The signal-to-interference-plus-noise ratios (SINRs) are described by the ratio between the energy of the deterministic MPCs to the interfering DM plus noise
\begin{equation}
  \SINR_{k,n}^{(j)} = \frac{|\alpha_{k,n}^{(j)}|^2}{N_0 + \Tp
  S_{\nu,n}^{(j)}(\tau_{k,n}^{(j)}) }
 \label{eq:sinr}
\end{equation}
The according MPC range uncertainties $\sigma_{k,n}^{(j)\ist2} = \var{z_{k,n}^{(j)}}$ to already associated VAs is given as
\begin{equation}
   \sigma_{k,n}^{(j)\ist2} \geq \left(\frac{8\pi^2\beta^2}{c^2} \SINR_{k,n}^{(j)} \right)^{-1}.
  \label{eq:SINRrange}
\end{equation}

\subsection{Geometry-based Probabilitstic Environment Model (GPEM)}\label{ch2_ProbFP}

\begin{figure}[ht]
		\psfrag{pmiss}{ \textcolor{black}{\raisebox{.1em}{\hspace{-.2em} $\vm{a}_{\mathrm{false}}$}}}
		\psfrag{p1j}{ \textcolor{black}{\raisebox{-.4em}{\hspace{-1.2em} $\vm{a}_1^{(j)}$}}}
		\psfrag{p2j}{ \textcolor{black}{\raisebox{0em}{\hspace{-.7em} $\vm{a}_2^{(j)}$}}}
		\psfrag{p3j}{ \textcolor{black}{\raisebox{0em}{\hspace{-1.2em} $\vm{a}_3^{(j)}$}}}
		\psfrag{p4j}{ \textcolor{black}{\raisebox{0em}{\hspace{-1.2em} $\vm{a}_4^{(j)}$}}}
		\psfrag{pm}{ \textcolor{black}{\raisebox{0em}{\hspace{-.6em} $\vm{p}$}}}
  \centering
  \includegraphics[width=1\columnwidth]{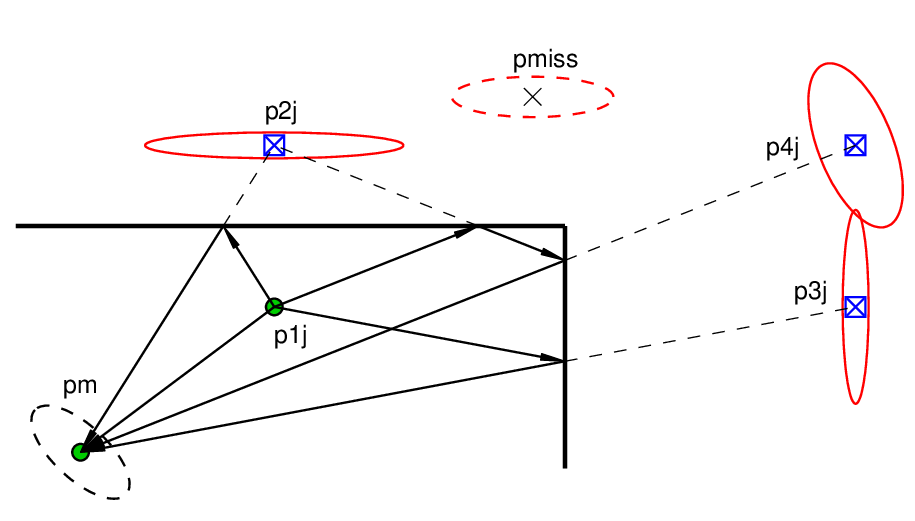}
  \caption{Illustration of the VA for the $j$-th anchor and an agent with PDF $p\big(\vm{a}\mpcidx^{(j)}\big)$ and $p\big(\vm{p}\agentidx\big)$, respectively. The VA at position $\vm{a}_{\mathrm{false}}$ represents a false detected VA.} 
  \label{fig:prob_VAgeometry}
\end{figure}

Fig.~\ref{fig:prob_VAgeometry} illustrates the probabilistic geometric environment model. A signal exchanged between an anchor at position $\vm{a}_1^{(j)}$ and an agent at $\vm{p}\agentidx$ contains specular reflections at the room walls, indicated by the black lines\footnote{Since the radio channel is reciprocal, the assignment of transmitter and receiver roles to anchors and agents is arbitrary and this choice can be made according to the application scenario.}. These reflections can be modeled geometrically using the VA $\vm{a}\mpcidx^{(j)}$ with $k = 1,\dots,K^{(j)}$ that are mirror-images of the $j$-th anchor w.r.t.\ walls \cite{MeissnerPhD2014,Borish1984,KunischICUWB2003}. The number of VAs per anchor $j$ is defined as $K^{(j)}$. The VAs of all anchors are comprised in $\VA{n} = \big\{ \VA{n}^{(j)} \big\}^J_{j =1}$, where $\VA{n}^{(j)} = \big\{ \vm{a}_{k,n}^{(j)} \big\}_{k = 1}^{K_n^{(j)}}$. To be able to cope with uncertainties in the floor plan the deterministic geometric model of the VA positions $\vm{a}\mpcidx^{(j)}$ of the $j$-th anchor, is extended to a probabilistic one as shown in Fig.~\ref{fig:prob_VAgeometry}. The VA positions and the agent position $\vm{p}\agentidx$ are represented by a joint PDF $p\big(\vm{p}\agentidx, \vm{a}_1^{(j)}, \vm{a}_2^{(j)}, \dots, \vm{a}_{K_n^{(j)}}^{(j)}\big)$. If the position of the $j$-th anchor is assumed to be known exactly, the joint PDF reduces to $p\big(\vm{p}\agentidx, \vm{a}_2^{(j)}, \dots, \vm{a}_{K_n^{(j)}}^{(j)}\big)$. 

%geometry explanation
%%%%%%%%%%%%%%%%%%%%%%%%%%%%%%%%%%%%%%%%%%%%%%%%%%%%%%%%%%%%%%%%%%
%VA figure
%%%%%%%%%%%%%%%%%%%%%%%%%%%%%%%%%%%%%%%%%%%%%%%%%%%%%%%%%%%%%%%%%%

The joint PDF of the agent and the VA positions is represented by a multivariate Gaussian RV, where the figure shows the marginal distributions of the agent $p\big(\vm{p}\agentidx\big)$ (dashed black ellipses) and the VA positions $p\big(\vm{a}\mpcidx^{(j)})$ (red ellipses). The marginal distribution $p\big(\vm{a}_{\mathrm{false}}\big)$ (dashed red ellipse) defines a wrongly detected VA at position $\vm{a}_{\mathrm{false}}$. The anchor position $\vm{a}_1^{(j)}$ is assumed to be known perfectly. Uncertainty in the floor plan does not just mean that the VA positions are uncertain and thus described by RV, but also that floor plan information is incorrect/inconsistent or entirely missing. This means that positioning and tracking algorithms based on VA, have to consider this lack of knowledge.

\subsection{Probabilistic Data Association (PDA)} \label{subsec:dataassoc}

The state of the agent $\vm{x}_n = [\vm{p}_n\trans,\vm{v}_n\trans]\trans$, where $\vm{v}_n$ is the velocity, evolves according to the state transition probability density function (PDF) $p( \vm{x}_n | \vm{x}_{n-1})$ over time instances $n$. From each VA in $\VA{n}^{(j)}$ and the predicted agent position, a set of expected MPC distances $\D{n}^{(j)}$ at time step $n$ is computed. The MPC distances described in Section~\ref{sec:channelest} are subject to a data association uncertainty, i.e., it is not known which measurement in $\vm{z}_n^{(j)}$ originated from which VA $k$ of the $j$-th anchor, and it is also possible that a measurement $y_{m,n}^{(j)}$ did not originate from any VA (false alarm, clutter) or that a VA did not give rise to any measurement (missed detection). The probability that a VA is detected is denoted by $\rmv P_{\text{d}}$. Possible associations at time instance $n$ are described by the $K_n^{(j)}$-dimensional random vector $\vm{b}_n^{(j)} = \big[b_{1,n}^{(j)} \cdots b_{n,K_n^{(j)}}^{(j)} \big]\trans$, whose $k$-th entry is defined as \cite{LeitingerGNSS2016,LeitingerICC2017,LeitingerTWC2019,LeitingerICC2019,MeyerProc2018,MeyerIFC2015}
\[ 
b_{k,n}^{(j)} = \begin{cases} 
    m \in \{1,\dots,M\} \ist , & \begin{minipage}[t]{40mm}$\vm{a}\mpcidx^{(j)}$ generates
				       measurement $z_{m,n}^{(j)}$\end{minipage}\\[3.8mm]
   0 \ist, & \begin{minipage}[t]{40mm} $\vm{a}\mpcidx^{(j)}$ did not give rise
				      to any measurement.\end{minipage}
  \end{cases}
\]
We also define $\vm{b}_n = \big[\vm{b}_n^{(1)\ist\text{T}} \cdots\vm{b}_n^{(J)\ist\text{T}} \big]\trans$. False alarms are modeled by a uniform distribution with mean arrival rate $\mu$, and the distribution of each false alarm measurement is described by the PDF $f_{\text{FA}}\big( z_{m,n}^{(j)} \big)$ \cite{barShalom95,vermaak05}, factoring in a likelihood that a measurement correspond to a false alarm. 

The statistical dependence of the distance measurement vectors $\vm{z}_n = \big[\vm{z}_n^{(1)}, \cdots,  \vm{z}_n^{(J)} \big]\trans$ on the agent state vector $\vm{x}_n$ and the association vector $\vm{b}_n$ is described by the \emph{global likelihood function} $f( \vm{y}_n | \vm{x}_n, \vm{b}_n)$. Under commonly-used assumptions about the statistics of the measurements \cite{barShalom95,vermaak05,MeissnerPhD2014}, the global likelihood function at time instances $n$ factors as
\begin{align*}
	f( \vm{z}_n | \vm{x}_n, \vm{b}_n) &\ist=\ist \prod^{J}_{j = 1} \rmv\Bigg( \prod^{M}_{m = 1} \rmv\rmv f_{\text{FA}}\big( z_{m,n}^{(j)} \big) \rmv\Bigg) \\
	&\qquad \times \prod_{k \in \cl{Q}(\vm{x}_n,\vm{b}_n^{(j)})}\!\! \frac{f\Big(z_{b_{k,n}^{(j)},n}^{(j)} \rmv\Big|\ist \vm{x}_n; \vm{a}\mpcidx^{(j)}, \sigma_{k,n}^{(j)}\Big)}{f_{\text{FA}}\Big( z_{b_{k,n}^{(j)},n}^{(j)} \Big)} \,, 
  \nn
\end{align*}
where $\cl{Q}(\vm{x}_n,\vm{b}_n^{(j)}) \triangleq \big\{k \rmv\in\rmv \{1,\dots,K_n^{(j)}\} : b_{k,n}^{(j)} \!\neq\rmv 0, \big\}$. The local likelihood function $ f\big( z_{m,n}^{(j)} | \vm{x}_n; \vm{a}\mpcidx^{(j)}, \sigma_{k,n}^{(j)} \big)$ is related to a noisy measurement of the distance to VA $\vm{a}\mpcidx^{(j)}$ at agent position $\vm{p}_n$ which is modeled as
\begin{equation}
	\label{eq:messmodel}
	z_{k,n}^{(j)} = \|\vm{p}_n - \vm{a}\mpcidx^{(j)}\| + v_{k,n}^{(j)} \,, \nn
\end{equation}
where $v_{k,n}$ is a zero-mean Gaussian random variable with standard deviation $\sigma_{k,n}^{(j)}$ as described in \eqref{eq:SINRrange}.
Based on the factorized likelihood model, a probabilistic data association algorithm is used to compute the associations between the expected delay to the VAs and the estimated MPCs using belief propagation as described in \cite{LeitingerGNSS2016,LeitingerICC2017,LeitingerTWC2019,LeitingerICC2019,MeyerProc2018,MeyerIFC2015}. The most probable MPC-to-anchor associations are obtained by means of an approximation of the maximum a posterior (MAP) detector \cite{kay1998}
\begin{equation}
  \hat{b}_{k,n}^{(j)\ist\text{MAP}} \triangleq \underset{b_{k,n}^{(j)} \in \{1,\dots, M\}}{\arg\max} p\big(b_{k,n}^{(j)}\big| \bd{z}\big).
  \label{eq:map}
\end{equation}
After the PDA was applied for all anchors, the following union sets are defined:
\begin{itemize}
  \item The set of associated discovered (and optionally a-priori known) VAs $\VAassoc{n} = \bigcup_j \VAassoc{n}^{(j)}$.
  \item The according set of associated measurements $\Z{n,\mathrm{ass}} = \bigcup_j \Z{n,\mathrm{ass}}^{(j)}$.
  \item The set of remaining measurements  $\Z{n,\overline{\mathrm{ass}}} = \bigcup_j \Z{n,\overline{\mathrm{ass}}}^{(j)}$,
  which are not associated to VAs of $\VAassoc{n}$.
\end{itemize}

\subsection{MINT-SLAM}\label{sec:BSSE}

In the most generic form, the prediction equation for the VAs $\VA{n}$ and the agent state $\vm{x}_n =  [\vm{p}_n, \vm{x}_n ]\trans$, can be written as, using the Markovian assumption,   
\begin{align}\label{eq:gen_pred_step_SLAM}
  p(\vm{x}_{n},\VA{n}|\Z{1:n-1}) &= \int p(\vm{x}_{n-1},\VA{n-1}|\Z{1:n-1}) p(\vm{x}_n|\vm{x}_{n-1})\nonumber \\
		&\quad \times  p(\VA{n}|\VA{n-1}) {d}\{\vm{x}_{n-1},\VA{n-1}\},
\end{align}
where $p(\vm{x}_n|\vm{x}_{n-1})$ and $p(\VA{n}|\VA{n-1})$ are the state transition probability distribution functions of the agent and the VAs, respectively. The latter can be represented by an identity function. The update equation is then
\begin{equation}\label{eq:gen_update_step_SLAM}
  p(\vm{x}_{n},\VA{n}|\Z{1:n}) =
\frac{p(\Z{n}|\vm{x}_{n},\VA{n})p(\vm{x}_n,\VA{n}|\Z{1:n-1
})}{p(\Z{1:n}|\Z{ 1:n-1})}, 
\end{equation}
where $p(\Z{n}|\vm{x}_{n},\VA{n})$ is the likelihood function of the current measurements. Assuming that the agent moves along a path according to a linear Gaussian constant-velocity motion, the state space model is defined as, 
\begin{align}\label{eq:ssmodel}
    \vm{x}_{n} & = \vm{F}\vm{x}_{n-1} + \vm{G}\vm{n}_{\mathrm{a},n} \nonumber \\
    &= 
     \begin{bmatrix}
       1 & 0 & \DT & 0 \\
       0 & 1 & 0 & \DT \\
       0 & 0 & 1 & 0\\\
       0 & 0 & 0 & 1
     \end{bmatrix}
    \vm{x}_n +
    \begin{bmatrix}
      \frac{\DT^2}{2} & 0 \\
      0 & \frac{\DT^2}{2} \\
      \DT & 0 \\
      0 & \DT \\
    \end{bmatrix}
 \vm{n}_{\mathrm{a},n},
\end{align}
where $\DT$ is the discrete time update rate. The driving acceleration noise term $\vm{n}_{\mathrm{a},n}$ is zero-mean, circular symmetric with variance $\sigma_\mathrm{a}^2$, and models motion changes which deviate from the constant-velocity assumption. The transformed noise covariance matrix is given as $\vm{R}_\mathrm{a} = \sigma_\mathrm{a}^2\vm{G}\vm{G}\trans$.  The entire state space of $\vm{x}_n$ and the associated VAs $\VAassoc{n}$ described in \eqref{eq:gen_pred_step_SLAM} are formulated as \cite{LeitingerICC2015, LeitingerPhD2016} 
\begin{equation}\label{eq:ssmodel_SLAM}
		\tilde{\vm{x}}_{n} = 
		\begin{bmatrix}
				\vm{F} & \vm{0}_{4 \times 2K_n} \\
				\vm{0}_{2K_n \times 4} & \vm{I}_{2K_n \times 2K_n} \\
		\end{bmatrix}
		\tilde{\vm{x}}_{n-1} +
		\begin{bmatrix}
				\vm{G} \\
				\vm{0}_{2K_n \times 2}
		\end{bmatrix}
		\vm{n}_{\mathrm{a},n},
\end{equation}
where $\tilde{\vm{x}}_n =[\vm{x}_n\trans,\vm{a}_{2,n}\trans,\dots,\vm{a}_{K_n,n}\trans]\trans$ represents the stacked state vector with $\{ \vm{a}_{k,n}^{(j)} \} \in \VAassoc{n}$. The covariance matrix of the state vector consists of the agent covariance matrix $ \vm{C}_{\vm{x}_n}$, the cross-covariances $\vm{C}_{\vm{x}_n, \vm{a}_{k,n}}$ between the agent state $\vm{x}_n$ and the VAs at positions $\vm{a}_{k,n}$, the cross-covariances between VAs $\vm{C}_{\vm{a}_{k,n},\vm{a}_{k',n}}$ with $k \neq k'$, and the covariances of the VAs $\vm{C}_{\vm{a}_{k,n}}$. The measurement model is defined as
\begin{align}\label{eq:msmodel_SLAM}
  \vm{z}_n = \tilde{\vm{h}}_n(\tilde{\vm{x}}_n) + \tilde{\vm{n}}_{z,n},
\end{align}
where $\vm{z}_n$ is defined in \eqref{eq:messmodel} with the according stack measurement noise vector $\tilde{\vm{n}}_{z,n}$. The measurement model $\tilde{\vm{h}}_n$ contains all distance equations $||\vm{a}_{n,k}^{(j)} -\vm{p}_n|| \; \forall\: \vm{a}_{n,k}^{(j)} \in \VAassoc{n}$ to update the agent and the VAs, respectively. As Bayesian state estimator a UKF is used \cite{LeitingerPhD2016}. The measurement covariance matrix is written as
\begin{equation}
  \vm{R}_n =\mathrm{diag} \left\{ \var{z_{k, n}^{(j)}} \right\}  \;\; \forall
\:k,j: \vm{a}_{k,n}^{(j)} \in \VAassoc{n}, \label{eq:covar_measmodel}
\end{equation}
where the range variances are defined by \eqref{eq:SINRrange}.

%%%%%%%%%%%%%%%%%%%%%%%%%%%%%%%%%%%%%%%%%%%%%%%%%%%%%%%%%%%%%%%%%%%%%%%%%%%%%%%%%%%%%%%%%%%%%%%%%%%%%
%%%%%%%%%%%%%%%%%%%%%%%%%%%%%%%%%%%%%%%%%%%%%%%%%%%%%%%%%%%%%%%%%%%%%%%%%%%%%%%%%%%%%%%%%%%%%%%%%%%%%

\section{Cognitive Positioning System}
\label{sec:coglocmodel}
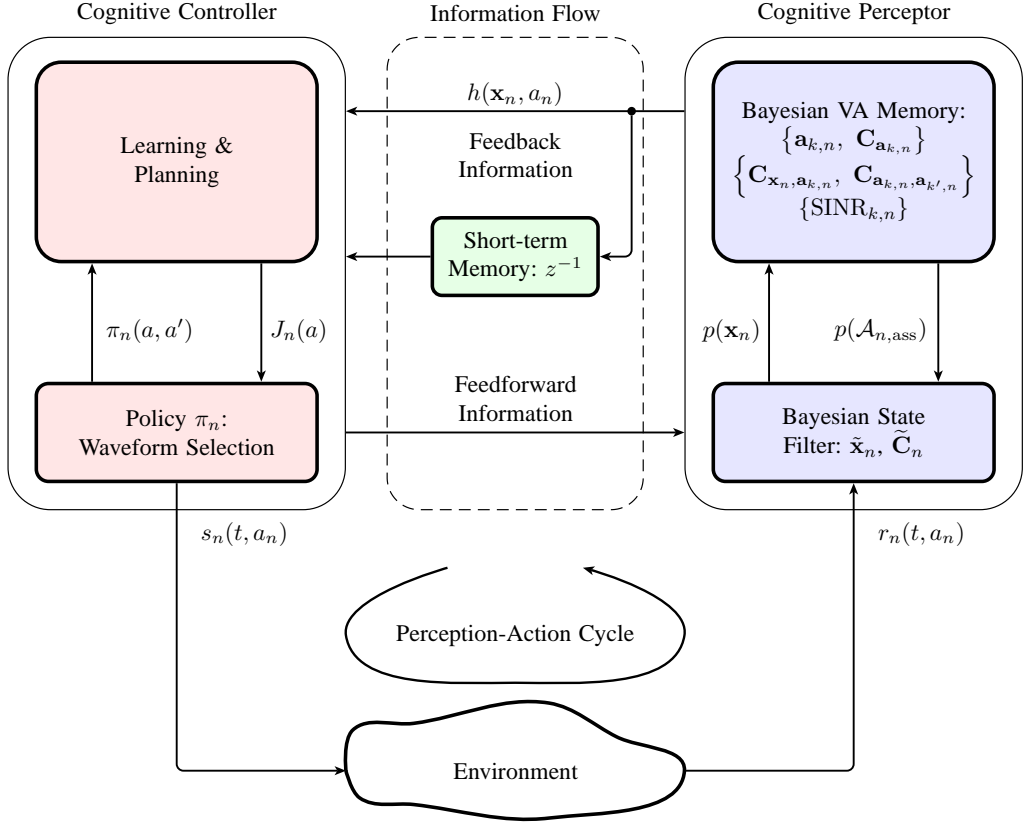
\begin{figure*}[t!]
	\normalsize
	\vspace{3mm}
	\centering
	\scalebox{.9}{
	\begin{pspicture}[showgrid=false](-3,-2)(11,11)
		
		\pssignal(-.5,2.5){s}{$s\timestep(t,a\timestep)$}
  \pssignal(9.5,2.5){r}{$r\timestep(t,a\timestep)$}
  \pssignal(0.3,5.5){pi}{$J\timestep(a)$}
  \pssignal(-1.9,5.5){CtGf}{$\pi\timestep(a,a')$}

  \pssignal(6.7,5.5){CtGf}{$p(\vm{x}\timestep)$}
  \pssignal(8.9,5.5){pi}{$p(\VAassoc{\timestepsym})$}

  \pssignal(-1.5,10.2){CC}{Cognitive Controller}
  \pssignal(8.5,10.2){CP}{Cognitive Perceptor}
		\pssignal(3.5,10.2){IF}{Information Flow}
  \pssignal(3.5,4.5){FF}{\parbox[c]{4\psunit}{\centering Feedforward\\Information}}
  \pssignal(3.5,8.1){FB}{\parbox[c]{4\psunit}{\centering Feedback\\Information}}

  \pssignal(3.5,9){Entropy}{\parbox[c]{4\psunit}{\centering $h(\vm{x}\timestep, a\timestep)$}}

		% System connection
		%%%%%%%%%%%%%%%%%%%%%%%%%%%%%%%%%%%%%%%%%%%%%%%%%%%%%%%%%%%%%%%%%%%%%%%%%%%%%%%%%%%%
		%%%%%%%%%%%%%%%%%%%%%%%%%%%%%%%%%%%%%%%%%%%%%%%%%%%%%%%%%%%%%%%%%%%%%%%%%%%%%%%%%%%%
		\rput(7.25,4.75){\rnode{perc_mem_out}{}}
		\rput(7.25,6.5){\rnode{perc_mem_in}{}}
  %%%%%%%%%%%%%%%%%%%%%%%%%%%%%%%%%%%%%%%%%%%%%%%%%%%%%%%%%%%%%%%%%%%%%%%%%%%%%%%%%%%%
		\rput(9.75,4.75){\rnode{mem_perc_in}{}}
		\rput(9.75,6.5){\rnode{mem_perc_out}{}}

  %%%%%%%%%%%%%%%%%%%%%%%%%%%%%%%%%%%%%%%%%%%%%%%%%%%%%%%%%%%%%%%%%%%%%%%%%%%%%%%%%%%%
		\rput(-2.75,4.75){\rnode{action_mem_out}{}}
		\rput(-2.75,6.5){\rnode{action_mem_in}{}}
  %%%%%%%%%%%%%%%%%%%%%%%%%%%%%%%%%%%%%%%%%%%%%%%%%%%%%%%%%%%%%%%%%%%%%%%%%%%%%%%%%%%%
		\rput(-.25,4.75){\rnode{mem_action_in}{}}
		\rput(-.25,6.5){\rnode{mem_action_out}{}}

  %%%%%%%%%%%%%%%%%%%%%%%%%%%%%%%%%%%%%%%%%%%%%%%%%%%%%%%%%%%%%%%%%%%%%%%%%%%%%%%%%%%%
		\rput(6,8.75){\rnode{Pmen_Amem_out}{}}
		\rput(1,8.75){\rnode{Pmen_Amem_in}{}}
  %%%%%%%%%%%%%%%%%%%%%%%%%%%%%%%%%%%%%%%%%%%%%%%%%%%%%%%%%%%%%%%%%%%%%%%%%%%%%%%%%%%%
		\dotnode(5.22,8.75){Pmen_Smem_out}
		\rput(1,6.6){\rnode{Amen_Smem_in}{}}
  %%%%%%%%%%%%%%%%%%%%%%%%%%%%%%%%%%%%%%%%%%%%%%%%%%%%%%%%%%%%%%%%%%%%%%%%%%%%%%%%%%%%
		\rput(1,4){\rnode{action_perc_out}{}}
		\rput(6,4){\rnode{action_perc_in}{}}

		%Blocks
		%%%%%%%%%%%%%%%%%%%%%%%%%%%%%%%%%%%%%%%%%%%%%%%%%%%%%%%%%%%%%%%%%%%%%%%%%%%%%%%%%%%%
  %%%%%%%%%%%%%%%%%%%%%%%%%%%%%%%%%%%%%%%%%%%%%%%%%%%%%%%%%%%%%%%%%%%%%%%%%%%%%%%%%%%%
		\psset{style=RoundCorners,gratioWh =1.25}

		% Action			
		\psfblock[linewidth=1.5pt,framesize=4.2 1.5,FillColor=red!10](-1.5,4){Actor}{\parbox[c]{4\psunit}{\centering Policy $\pi\timestep$:\\Waveform Selection}}
		\psfblock[linewidth=1.5pt,framesize=4.2 3, FillColor=red!10](-1.5,8){A_Mem}{\parbox[c]{2.8\psunit}{\centering Learning \& Planning}}
		\psfblock[linewidth=.5pt,framesize= 5 7](-1.5,6.35){box_actor}{}

		% Perception

		\psfblock[linewidth=1.5pt,framesize=4.2 1.5,FillColor=blue!10](8.5,4){Perceptor}{\parbox[c]{3\psunit}{\centering Bayesian State Filter: $\tilde{\vm{x}}\timestep$, $\widetilde{\vm{C}}\timestep$}}
		\psfblock[linewidth=1.5pt,framesize=4.2 3,FillColor=blue!10](8.5,8){P_Mem}{\parbox[c]{3.8\psunit}{\centering Bayesian VA Memory: \\ $\left\{\vm{a}_{\mpcidxsym,\timestepsym},\ \vm{C}_{\vm{a}_{\mpcidxsym,\timestepsym}}\right\}$ \\ $\left\{\vm{C}_{\vm{x}\timestep,\vm{a}_{\mpcidxsym,\timestepsym}}, \ \vm{C}_{\vm{a}_{\mpcidxsym,\timestepsym},\vm{a}_{\mpcidxsym',\timestepsym}}\right\}$ \\ $\left\{\SINR_{\mpcidxsym,\timestepsym}\right\}$}}

  \psfblock[linewidth=.5pt,framesize= 5 7](8.5,6.35){box_perceptor}{}

		% Working Memory
		\psfblock[linewidth=1.5pt,framesize= 2.5 1.2, FillColor=green!10](3.5,6.6){SMem}{\parbox[c]{2.5\psunit}{\centering Short-term\\Memory: $z^{-1}$}}
		\psfblock[linewidth=.5pt,linestyle=dashed,framesize= 3.8 7](3.5,6.35){W_Mem}{\parbox[c]{2\psunit}{}}

		% Environment
		\rput(6,-1){\rnode{environ_output}{}}
  \rput(1,-1){\rnode{environ_input}{}}
		\rput(3.5,-1){Environment}
		\pscurve[linewidth=1.5pt,  showpoints=false](1,-1)(1.2,-1.4)(2,-1.5)(4,-1.7)(6,-1)(5,-.5)(4,0)(2,-.3)(1.1,-.4)(1,-1)

		\psset{style=Arrow}
		% PAC
		\pscurve[linewidth=1pt,showpoints=false](2.5,2)(1,1)(3.5,.3)(6,1)(4.5,2)
		\rput(3.5,1){Perception-Action Cycle}

		%Connections
  %%%%%%%%%%%%%%%%%%%%%%%%%%%%%%%%%%%%%%%%%%%%%%%%%%%%%%%%%%%%%%%%%%%%%%%%%%%%%%%%%%%%
  %%%%%%%%%%%%%%%%%%%%%%%%%%%%%%%%%%%%%%%%%%%%%%%%%%%%%%%%%%%%%%%%%%%%%%%%%%%%%%%%%%%%
%
		\ncangle[angleA=180, angleB=-90]{environ_output}{Perceptor}
		\ncangle[angleA=-90, angleB=0]{Actor}{environ_input}
		\ncangle[angleA=90, angleB=-90]{perc_mem_out}{perc_mem_in}
  \ncangle[angleA=-90, angleB=90]{mem_perc_out}{mem_perc_in}
		\ncangle[angleA=90, angleB=-90]{action_mem_out}{action_mem_in}
  \ncangle[angleA=-90, angleB=90]{mem_action_out}{mem_action_in}

		\ncangle[angleA=180, angleB=0]{Pmen_Amem_out}{Pmen_Amem_in}
  \ncangle[angleA=-90, angleB=0]{Pmen_Smem_out}{SMem}
  \ncline[angleA=180, angleB=0]{SMem}{Amen_Smem_in}
  \ncangle[]{action_perc_out}{action_perc_in}

	\end{pspicture}}
	\caption{Block diagram of the cognitive indoor positioning and tracking system that uses multipath channel information.}
	\label{fig:CL_Blockdiag}
% \vspace*{2mm}
\end{figure*}

The basic building blocks of a CDS, namely the \textit{perception-action cycle (PAC)}, \textit{cognitive perceptor (CP)},  \textit{information feedback} and the \textit{cognitive controller (CC)} are depicted in Fig.~\ref{fig:CL_Blockdiag}. All of these blocks are reciprocally coupled and form a hierarchical structure to enable the ability to interpret the environmental observables on different abstraction layers.

\subsection{Multipath-assisted Positioning as {CDS}}\label{sec:buildingblocks}

Figure~\ref{fig:CL_Blockdiag} illustrates the block diagram of a cognitive localization and tracking system with a triple layered structure:
\begin{itemize}
	\item \emph{First Layer}: Defines (i) the direct Bayesian state estimation $p\big( \vm{x}_n \big|\Z{n}, \vm{c}_n \big)$ at the {CP} holds the agent position and its velocity, and (ii) the cognitive control parameters $\vm{c}_n$ at the {CC} based on the feedback information of the Bayesian state space filter.   
	\item \emph{Second Layer}: Represents (i) the memory for the {GPEM} described by the VAs with marginal {PDF} $p\big(\VA{n}^{(j)}|\Z{n}^{(j)}, \vm{c}_n \big)$ and the memory for the {GSCM} described by the $\SINR_{k,n}^{(j)}$ of the {MPC} at the {CP} and (ii) the memory of VAs specific waveform parameters at the {CC}, which specify on which the cognitive control is based on.
	\item \emph{Third Layer}: It represents the highest layer and is different from the two layers below in the sense that it defines the application driven by the cognitive localization/tracking system. The {CP} memory of applications holds abstract parameters or structures of the specified application and the {CC} enables the motor control for realizing higher goal planning \cite{WymeerschICC2013}.
\end{itemize}
The first and second layers describe the signal and information processing of the model parameters of the surrounding physical environment and the radio channel. On the other hand, the third layer holds higher goal parameters, i.e. motor-control input to fulfill navigation goals, that are based on the physical-related parameters \cite{WymeerschICC2013, MeyerJSAC2015, GrocholskyPhD2002}.  

\subsection{Feedback Information}\label{sec:feedback}

The system is able to adapt online its behavior to the environment, i.e. perceptual attention is given, through the following principles:
\begin{itemize}
	\item At the CP side, the GSCM and GPEM memories are updated using the received signal $r_n(t,\vm{c}_n)$ with waveform parameters chosen by the CC. 
	\item In the actual sensing cycle the attention is put through the CC using the control parameters $\vm{c}_n$ on the potential set of VA and their parameters memorized in the GSCM and GPEM. These model parameters are seen at the CP side of Fig.~\ref{fig:CL_Blockdiag}.
\end{itemize}
Now the question is, ``\emph{How to control the environment information flow through the received signal and put cognitive attention on the relevant features in the following sensing cycle?}'' The answer to this lies in the {CC} and the feed-back and feed-forward information between the perceptor and the controller as illustrated in Fig.~\ref{fig:CL_Blockdiag}.

The control parameter vector $\vm{c}_{n+1}$ of the next sensing cycle is chosen in order to gain the most ``valuable'' position-related information from the new set of measurements $\tilde{\mathrm{Z}}_{n+l}$ using the predicted posterior $p\big( \vm{x}_{n+l}, \VA{n+l} | \tilde{\mathrm{Z}}_{n+l}, \vm{b}_{n+l}, \vm{c}_{n+l} \big)$ the predicted received signals $\tilde{\mathrm{Z}}_{n+l}$ that depends on the chosen signal model, with $l = 0,\dots,l_\mathrm{future}$ as future horizon. 
This goal can be reached by minimizing an expected cost-to-go function, yielding
\begin{equation}\label{eq:optmizeUtility}
 \vm{c}_{n+1} = \underset{\vm{c}_n}{\arg \min}\quad \mathcal{C}\big(p\big( \vm{x}_{n+l}, \VA{n+l} | \tilde{\mathrm{Z}}_{n+l},\vm{b}_{n+l}, \vm{c}_{n+l} \big)\big), 
\end{equation} 
where $\mathcal{C}(\cdot)$ is the expected cost-to-go function for optimal control \cite{GrocholskyPhD2002, Chaloner1995} of the environmental information contained in $\tilde{\mathrm{Z}}_{n+l}$. The expected cost-to-go function is based on an information-theoretic measure that should depend on the environment parameters, like the {VA} specific $\SINR_{k,n}^{(j)}$, and serves as feedback information in the {CDS}. 

In general, estimation and control problems have to deal with probabilistic states and observations. As a consequence, also the control has to be probabilistic, i.e.\ the cost function or utility must handle uncertainties. Based on covering the uncertainty of the state with a {PDF}, a measure of informativeness of measurements has to be defined on the posterior state distribution. Two commonly used information measures of an {RV} are the entropy \cite{Shannon1948} and the Fisher information \cite{Kay1993}.
\subsection{Information Measures for Feedback}\label{sec:ch2_informationmeas}

\subsubsection{Fisher Information}\label{sec:FI}

The Fisher information matrix (FIM) of a RV $\vm{r}$, dependent on the deterministic parameter $\vm{p}$, can also be used as a measure of information. Using the likelihood function $\ln f(\vm{r};\vm{p} ) $, it is defined as 
\begin{equation}
  \FIM_\vm{p} = \E{\vm{r};\vm{p} }
  {\left[\frac{\partial}{\partial\vm{p} } \ln f(\vm{r};\vm{p} ) \right]
  \left[\frac{\partial}{\partial\vm{p} } \ln f(\vm{r};\vm{p} ) \right]\trans}. 
\end{equation}

\subsubsection{Entropy}\label{sec:ch2_entropy}

For a continuous-valued vector {RV} $\vm{p} \in \mathbb{R}^L$ (in the follow-up sections $\vm{p}$ represents the agent position), the conditional entropy is given as \cite{Cover2006}
\begin{equation}
 h(\vm{p}) \doteq -\E{\vm{p}}{\ln p(\vm{p})}  =-\int_{-\infty}^\infty\cdots \int_{-\infty}^\infty p(\vm{p})\ln p(\vm{p}) \mathrm{d}\vm{p},
\end{equation}
The entropy is directly related to the uncertainty of the according {RV}. For a multivariate Gaussian {RV} $\mathcal{N}\left(\vm{m}_\vm{p}, \vm{C}_\vm{p}\right)$ this means that the entropy is directly related to the covariance matrix $\vm{C}_\vm{p}$, yielding
\begin{equation}
  h(\vm{p}) = \frac{1}{2} \ln \left( (2\pi e)^L  \det\vm{C}_\vm{p}\big) \right),
\end{equation}
where $\det(\cdot)$ defines the determinant of a matrix. The determinate of the covariance matrix $\vm{C}_\vm{p}$ is a measure of the ``volume'' of uncertainty of $\vm{p}$. The more compact the volume is, the smaller is the entropy $h(\vm{p})$ and consequently the more informative is the distribution $p(\vm{p})$. 

The inverse of the FIM is a lower bound on the covariance $\vm{C}_{\hat{\vm{p}}} \succeq \FIM_\vm{p}^{-1}$ of the deterministic parameter $\vm{p}$ of an estimator $\hat{\vm{p}}$ \cite{Kay1993}. Looking at the entropy of the estimator's distribution $\mathcal{N}\left(\hat{\vm{p}}, \vm{C}_{\hat{\vm{p}}}\right)$, the explicit relationship between the FIM $\FIM_\vm{p}$ of $\vm{r}$ (dependent on $\vm{p}$) and the entropy $h(\hat{\vm{p}})$ is given as
\begin{align}\label{eq:relationentroFIM}
  h(\hat{\vm{p}}) &= \frac{1}{2} \log\left( (2\pi e)^L \det\big(\vm{C}_{\hat{\vm{p}}}\big) \right)  \\
		&\geq  -\frac{1}{2} \log\left( (2\pi e)^L \det\big(\FIM_\vm{p} \big)\right)\nonumber . 
\end{align}
As the relationship in \eqref{eq:relationentroFIM} shows, one can connect the FIM of a parameter vector with the entropy, resulting in a scalar measure of information that is valuable for choosing optimal waveform parameters, as it is needed for a cognitive positioning system. As it is shown in Section~\ref{sec:SINRest}, the FIM $\FIM_\vm{p}$ on the position of the agent $\vm{p}$ contains the environment and signal parameters, e.g. VA positions and the according SINRs. With this, a direct relationship between the environment, the feedback information and the control of the sensing is given, closing the PAC (Figure~\ref{fig:CL_Blockdiag}). In the same manner, the system can also be expanded to information-based control of the agent state to increase the informativeness in the measurements \cite{HoffmannTAC2010,JulianJRR2012,MeyerJSAC2015}.

%%%%%%%%%%%%%%%%%%%%%%%%%%%%%%%%%%%%%%%%%%%%%%%%%%%%%%%%%%%%%%%%%%%%%%%%%%%%%%%%%%%%%%%%%%%%%%%%%%%%%
%%%%%%%%%%%%%%%%%%%%%%%%%%%%%%%%%%%%%%%%%%%%%%%%%%%%%%%%%%%%%%%%%%%%%%%%%%%%%%%%%%%%%%%%%%%%%%%%%%%%%

\section{Cognitive MINT}
\label{sec:cogMINT}

\subsection{Cognitice Controller: Reinforcement Learning (RL)}\label{sec:RL}

As already stated in Section~\ref{sec:coglocmodel}, the control parameters should be chosen in order to optimize the expected cost-to-go function $\mathcal{C}\left( \cdot \right)$ of the predicted posterior {PDF} as defined in \eqref{eq:optmizeUtility}. In general, the expected cost-to-go function for a Bayesian state space filter can be written as 

\begin{equation}
 \mathcal{C}\left( p(\vm{x}_{n+1},\VA{n+1}|\tilde{\vm{r}}_{ n+1}(t,\vm{c}_n)\right) = \bar{g}\left( \boldsymbol{\epsilon}_{n+1|n+1}(\vm{c}_n)\right),  
\end{equation}
where $\boldsymbol{\epsilon}_{n+1|n+1}(\vm{c}_n)$ is the predicted posterior state-estimation error depend on the control parameters and $\bar{g}(\cdot)$ defines the cost-to-go function of the transmitter. The conditional entropy was discussed as a possible information measure for the feedback, thus a possible cost-to-go function $\bar{g}(\cdot)$ of the transmitter is the conditional entropy of the predicted posterior state-estimation error $\boldsymbol{\epsilon}_{n+1|n+1}(\vm{c}_n)$, given as $\bar{g}\left( \boldsymbol{\epsilon}_{n+1|n+1}(\vm{c}_n)\right) = h\left(\boldsymbol{\epsilon}_{n+1|n+1}(\vm{c}_n)\right)$ \cite{HaykinBook2012, Cover2006}. This entropy conditioned on the control parameter vector $\vm{c}_n$ is directly coupled with the posterior covariance matrix of the Bayesian tracking filter that is lower bounded by the inverse of the {EFIM} in \eqref{eq:EFIM_pos}. The entropy of the predicted posterior state-estimation error (when assuming a Gaussian approximation) is given as

\begin{align}\label{eq:entropstatetime}
 h\left( \boldsymbol{\epsilon}_{n+1|n+1}(\vm{c}_n) \right)  \propto \det\big( \widetilde{\vm{C}}_{\vm{x}_{n+1}}(\vm{c}_n) \big),
\end{align}
where $ \widetilde{\vm{C}}_{\vm{x}_{n+1}}(\vm{c}_n)$ and $\FIM_{\vm{x}_{n+1}}(\vm{c}_n)$ is the predicted state covariance matrix as described in Section \ref{sec:BSSE} of the state vector provided from the Bayesian state space filter (UKF) dependent on the control parameter vector $\vm{c}_n$. Thus, the entropy in \eqref{eq:entropstatetime} is directly coupled with the position-related information that is contained in the measurement noise covariance matrix $\vm{R}_{\mathrm{z},n}$ described by \eqref{eq:covar_measmodel}. How the introduced algorithm is using the state space and measurement model equations of the Bayesian state space estimator is described in more detail in Sections~\ref{sec:planning} and \ref{sec:WFL}.

For readability of the following derivations of the control optimization algorithm, the cost-to-go of the {CC} \eqref{eq:entropstatetime} is rewritten as $\bar{g}\left(\boldsymbol{\epsilon}_{n+1|n+1}(\vm{c}_n)\right) = h\left(\vm{x}_{n+1},\vm{c}_n\right)$ with $\vm{c}_n \ \in \mathcal{A}$, where $\mathcal{A}$ is the space of cognitive action with size $|\mathcal{A}|$ that represents the waveform library in our case. Consequently, the next set of waveform parameters has to be chosen in order to minimize the cost-to-go of the next posterior entropy. As elaborated in \cite{Bellman1957}, dynamic programming represents an optimal solution for such problems, but unfortunately it is based on the assumption that the state to be controlled is ``perfectly'' perceivable. Hence, methods have been introduced that are capable of handling imperfect state information \cite{Bertsekas2000} with the drawback that they are computational complex. In \cite{HaykinBook2012,HaykinPROC2012a} approximate dynamic programming was used for optimal control. In there, the trace of the posterior covariance matrix was used as cost-to-go function to reduce the computational complexity. The policy for control parameter selection in the transmitter at time instance $n$ is seeking to find the set of waveform parameters, for which the cost-to-go function $\bar{g}(\boldsymbol{\epsilon}_{n+1|n+1}(\vm{c}_n)) \approx \tr[\widetilde{ \vm{C}}_{\vm{x}_{n+1}}(\vm{c}_n)]$ is minimized for a rolling future horizon of $l_\mathrm{future}$ predicted states. In practice, it is difficult to construct all state transition probabilities from one state to another that are conditioned on the selected actions, including their cost incurred as a result of each transition. {RL}\footnote{{RL} represents an intermediate learning procedure that lies between supervised and unsupervised learning as stated in \cite{HaykinBook2012}.} \cite{Sutton1998} represents an approximation of dynamic programming \cite{Bellman1957, Bertsekas2000} for solving such optimal control and future planning task with high computationally efficiency. In {RL} literature the cost-to-go function is termed value-to-go function $J_n(\vm{c}_n)$ that is updated online for every {PAC} based on the immediate rewards $r_n$. The immediate reward $r_n$ is a measure of ``quality'' of an action $\vm{c}_n$ taken on the environment. Using the Markovian assumption and following the way in \cite{Fatemi2014}, it is given by                                                                                       
\begin{align}\label{eq:rewarddef}
 r_n =  g_n \left( h(\vm{x}_{n-1},\vm{c}_{n-1}) - h(\vm{x}_{n},\vm{c}_n)\right),
\end{align}
where $h(\vm{x}_{n},\vm{c}_n) \propto \det\big(\vm{C}_{\vm{x}_n}(\vm{c}_n)\big)$ and $g_n(\cdot)$ is an arbitrary scalar operator that in its most general form could also depend on the time instance $n$ \cite{Fatemi2014}. A reasonable function for the reward is the scaled change in the posterior entropy from one {PAC} to the next, i.e.
\begin{align}\label{eq:reward}
 r_n = \mathrm{sign}\left(\Delta h(\vm{x}_{n},\vm{c}_n\right)\left|\log\big(|\Delta h(\vm{x}_{n},\vm{c}_n)|\big)\right|. 
\end{align}
A positive reward will be favoring the current action $a_{n}$ for the future action $\vm{c}_{n+1}$ and conversely a negative one will lead to a penalty for these actions. As described in \cite{Fatemi2014}, the cognitive {RL} algorithm has to find the optimal future action $\vm{c}_{n+1}$ for the next {PAC} based on the immediate reward $r_n$ and the learned value-to-go function $J_n(\vm{c}_n)$.  

For computing the expected costs of future actions as it is done in dynamic programming, {RL} divides the computation of the value-to-go function into two parts, (i) the learning phase that incorporates the actual measured reward into the value-to-go function based on actions $\vm{c}_n$ and $\vm{c}_{n-1}$, and (ii) the planning phase that incorporates predicted future rewards into the value-to-go function. Whereas for learning a ``real'' reward is perceived from the environment, for planning just model-based predicted rewards are perceived from the internal perceptor memory using the feedforward link. A faster convergence to the optimal control policy can be achieved in this way.  

\subsection{Learning and Planning: Algorithm}\label{sec:learnplan}

The value-to-go function that is used in the cognitive controller is defined as \cite{Fatemi2014}
\begin{equation}
 J_n(\vm{c}) = \E{\pi_n}{r_n + \gamma r_{n+1} + \gamma^2 r_{n+2} + \cdots|\vm{c}_n = \vm{c}},
\end{equation}
where $r_n$ with $\quad \vm{c} \in \mathcal{C}$ is the actual reward, $r_{n+l}$ are the predicted future rewards that are based on the {GPEM} and {GSCM} parameter that are used by the Bayesian filter, $0 < \gamma \leq 1$ is the discount factor for future rewards based on action $\vm{c}_n \in \mathcal{C}$ and the expected value is calculated using the cognitive policy 
\begin{equation}
 \pi_n(\vm{c}',\vm{c}) = \mathbb{P}\left[\vm{c}_{n+1} = \vm{c}'|\vm{c}_{n} = \vm{c}\right], \quad \vm{c},\vm{c}' \in \mathcal{C},
\end{equation}
where $\mathbb{P}[\cdot|\cdot]$ defines a conditional {PMF} that describes the transition probabilities of all actions $\vm{c} \in \mathcal{C}$ over time instances $n$. Following the derivations in \cite{Fatemi2014}, the value-to-go function can be reformulated in an incremental recursive manner, yielding
\begin{equation}\label{eq:VtGrecursiveupdate}
 J_n(\vm{c}) \leftarrow J_n(\vm{c}) + \alpha \left[\mathcal{R}(\vm{c}) + \gamma \sum_{\vm{c}'}\pi_n(\vm{c},\vm{c}')J_n(\vm{c}') - J_n(\vm{c})\right],
\end{equation}
where $\mathcal{R}(\vm{c}) = \E{\pi_n}{r_n|\vm{c}_n = \vm{c}}$ $\forall \ \vm{c} \in \mathcal{C}$ denotes the expected immediate reward and $\alpha > 0$ is the learning rate. The algorithm for updating the value-to-go function can be found in the Appendix of \cite{LeitingerPhD2016}. The incremental recursive update in \eqref{eq:VtGrecursiveupdate} means that for all actions $\vm{c} \in \mathcal{C}$ the value-to-go function is updated using the expected immediate reward and the policy $\pi_n(\vm{c},\vm{c}')$ for all these actions.   

\subsubsection{Learning from applied Actions}\label{sec:learning}

With the value of the immediate reward $r_n$, a new value is learned for the value-to-go function for the currently selected action $\vm{c}_n$ using $J_n(\vm{c}_n) \leftarrow (1-\alpha)J_n(\vm{c}_n) + \alpha \mathcal{R}(\vm{c}_n)$ of \eqref{eq:VtGrecursiveupdate}. This accounts for the ``real'' physical action on the environment. Hence, only one parameter set can be chosen as an action for the {PAC} at a time; it would take at least $|\mathcal{C}|T$ seconds for applying all actions on the environment and collecting the according immediate rewards, where $T$ is the time period of a {PAC}. Unfortunately, this results in a poor convergence rate of the algorithm and unacceptable behavior for time-variant environments. A possible remedy against this is the planning of future actions based on the state space and measurement model of the Bayesian state estimator.   

\subsubsection{Planning for Improving Convergence Behavior}\label{sec:planning}

Planning is defined as predicting expected future rewards using the state and measurement model of the Bayesian state space filter to improve the convergence rate of the {RL} algorithm. As depicted in Fig.~\ref{fig:CL_Blockdiag}, the feedforward link is used to connect the controller with the perceptor. The feedforward information is a hypothesized future action, which is selected for a future planning stage. Inspecting \eqref{eq:VtGrecursiveupdate}, one can observe that for every action $\vm{c} \in \mathcal{M}$, where $\mathcal{M} \subset \mathcal{C}$ is a subset of $\mathcal{C}$ depending on the actual policy $\pi_n$, the predicted posterior covariance matrices $\widetilde{\vm{C}}_{n+l}(\vm{c})$ and the according predicted future rewards $r_{n+l}$, are computed with decreasing discount factor $\gamma^l$ for predicted future rewards, for $l = 1,\dots,l_\mathrm{future}$, where $l_\mathrm{future}$ is the future horizon. The predicted covariance matrices $\widetilde{\vm{C}}_{n+l}(\vm{c})$ for a specific future action $\vm{c}$ is computed using the state space (e.g. \eqref{eq:ssmodel_SLAM}) and measurement model (e.g.\ \eqref{eq:msmodel_SLAM}) of the Bayesian state space estimator and the according {GPEM} and {GSCM} parameters stored in the perceptors' memory as shown in Fig.~\ref{fig:CL_Blockdiag}. After the planning process is finished, the value-to-go function is updated for all actions $\vm{c} \in \mathcal{M}$. Finally, the actual {PAC} is closed by updating the policy to $\pi_{n+1}$ using the value-to-go function $J_{n+1}$ and choosing the new action, i.e.\ the waveform parameters, for the next {PAC} according to this new policy. This means that the value-to-go function $J_n(\vm{c}_n)$ and the policy $\pi_n$ are updated iteratively from one another from one {PAC} to the next {PAC}, with one important detail which is discussed below.

\paragraph{Explore/Exploit trade-off:} Both the planning process and choosing new actions are based on the policy. In planning, the chosen action-subset $\mathcal{M}$ is defined by sampling from the policy $\pi_n$ and new actions are selected based on the updated policy $\pi_{n+1}$. Hence, the policy is responsible for the explore/exploit trade-off in the action space. A widely used method for balancing the exploration of new actions and exploiting the already learned value-to-go function $J_n(\vm{c}_n)$ is the $\epsilon$-greedy strategy, meaning that with a small probability of $\epsilon$ a random action is selected, representing pure exploration, and with probability of $1-\epsilon$ the action is chosen according to the maximum of the value-to-go function, representing a pure exploitation. The random selection of a new action and the action in the subspace $\mathcal{M}$ can either be selected from a uniform distribution over the action space $\mathcal{A}$ or from the policy $\pi_n$. The policy is computed using the Boltzmann distribution
\begin{equation*}
	\pi_{n+l} = \pi_{n+l-1}(\vm{c})\frac{\exp\{\Delta J_{n+l}(\vm{c})/\tau\}}{\sum_{\vm{c}'}\pi_{n+l-1}(\vm{c}') \exp\{\Delta J_{n+l}(\vm{c}')/\tau\}},
\end{equation*}
where $\tau$ defines the exploration degree and is referred to as the system temperature \cite{Lazaric07reinforcementlearning} and $\Delta J_{n+l}(\vm{c}) = J_{n+l}(\vm{c}) - J_{n-1+l}(\vm{c})$. The cognitive action is selected according to
\begin{align}\label{eq:policy}
  \vm{c}_{n} = 
  \left\{ \begin{array}{ll}
    \text{random action} \sim \pi_{n+1} \in \mathcal{C} 
    & \ \text{if} \ \xi < \epsilon \\
    \arg\max_{\vm{c} \in \mathcal{C} } J_{n}(\vm{c})
    & \ \text{otherwise}
  \end{array} \right. \ ,
\end{align}
where $0 \leq \xi \leq 1$ is a uniform random number drawn at each time step $n$. As we have said, from the policy in \eqref{eq:policy} the new action $\vm{c}_{n+1}$ is selected and applied on the environment so that the next {PAC} can start. The important concept of \emph{attention} at the perceptor as well as the actuator side in the cognitive dynamic system can be argued with the following:
\begin{itemize}
 \item {\bf Perceptual attention:} Is given by the fact that the environment dependent parameters, i.e. the marginal {PDF} of the {VA} $p(\vm{a}_{k,n})$ and their multipath channel dependent reliability measures, $\SINR_{k,n}$, are learned and updated online, so that the perceptual Bayesian state space filter puts its attention on the relevant position-related information in the received signal.
  \item {\bf Control attention:} Is given by the fact that the policy $\pi_n$ that is learned over time and the according subset of actions $\mathcal{M}$ put focus on the ``more relevant'' actions. These action in turn focus on the relevant position-related information in the received signal. 
\end{itemize}
   
\subsubsection{Waveform Library}\label{sec:WFL}

The general form of the waveform library contains the control parameters $\vm{c}_n = \{ T_{\mathrm{p},n}, f_{\mathrm{c}, n}^j  \}_{j = 1}^J$ for the $j$-th anchor consisting of carrier frequencies and pulse durations. Hence, the {VA} specific {MPC} parameters are estimated using specific sub-bands of the radio channel spectrum defined by the parameter pair $T_{\mathrm{p},n}^j$ and $f_{\mathrm{c}, n}^j$, which in turn is chosen in an ``optimal'' manner. Optimal in this case means that the position-related information that is contained in the {MPC} parameters is maximized at agent position $\vm{p}_n$ (see for \eqref{eq:EFIM_pos}).   

Equations~\eqref{eq:sinr} and \eqref{eq:EFIM_pos}, which describe the parameters $\widetilde{\SINR}_{k,n}^j$, show the relation between the pulse parameter pair $T_{\mathrm{p},n}^j$ and $f_{\mathrm{c}, n}^j$ and the position-related information contained in the channel. The pulse duration $T_{\mathrm{p},n}^j$ scales the amount of {DM} and is directly proportional to the effective root mean square bandwidth $\beta$. The relation to $f_{\mathrm{c}, n}^j$ is not that obvious, since it describes the frequency dependency of the environment parameters and thus the {GSCM} parameters as the complex amplitudes of the {MPC} and the {DM} {PDP}. The set of selected {VA} should lead to the  highest overall {SINR} values (and accordingly the smallest range variances $\var{\hat{d}_{k,n}^j}$) and the smallest possible {GDOP}\footnote{The {GDOP} the ratio between position variance and the range variance \cite{Sahinoglu2008}. For positioning a small value indicates a high level of confidence that high precision can be reached. Hence, the {GDOP} indicates a ``good '' geometry for positioning, i.e. a good geometric placement of the anchors.}, i.e.\ geometric optimal constellation of {VA} positions which is reflected by the ranging direction matrix $\FIM_{\mathrm{r}}(\phi_{k,n}^j)$. In a cognitive sense this means that the actions $a \in \mathcal{A}$ are chosen to reduce the posterior entropy over time under quasi-stationary environment conditions.

%%%%%%%%%%%%%%%%%%%%%%%%%%%%%%%%%%%%%%%%%%%%%%%%%%%%%%%%%%%%%%%%%%%%%%%%%%%%%%%%%%%%%%%%%%%%%%%%%%%%%
%%%%%%%%%%%%%%%%%%%%%%%%%%%%%%%%%%%%%%%%%%%%%%%%%%%%%%%%%%%%%%%%%%%%%%%%%%%%%%%%%%%%%%%%%%%%%%%%%%%%%
\section{Results}
\label{sec:result}
\subsection{Measurement Setup}\label{ch7_setup}

\begin{figure}[t!]
  \centering
  \psfragoffset{xaxis}{\small}{black}{-.75}{0}{$x [m]$}
  \psfragoffset{yaxis}{\small}{black}{-.75}{.3}{$y [m]$}
  \psfragoffset{A1}{\small}{red}{-.2}{.2}{$\va{1}^{(1)}$}
		\psfragoffset{A2}{\small}{blue}{-.2}{.4}{$\va{1}^{(2)}$ }
		\includegraphics[width=1\columnwidth]{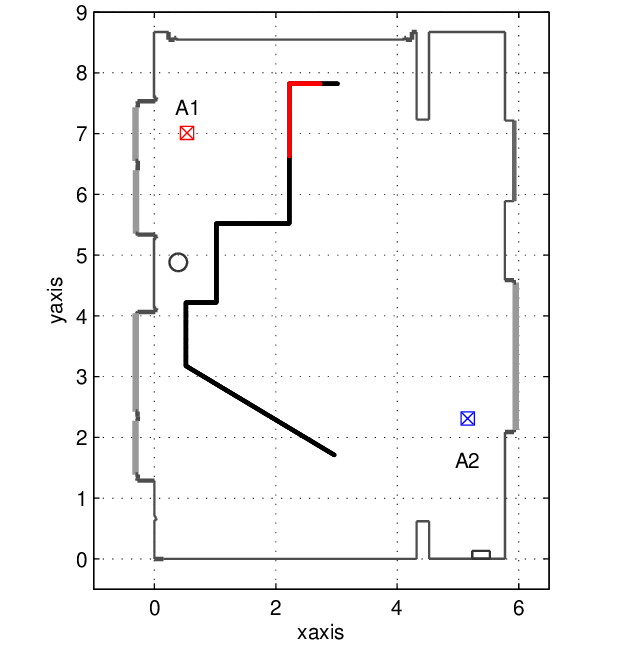}
  \caption{Scenario for probabilistic {MINT} using cognitive sensing in presence of additional {DM} interference. The anchors are at the positions $\vm{a}_1^{(1)}$ and $\vm{a}_1^{(2)}$. The black line represents the agent trajectory and the red part of the line indicates the agent positions, where the {DM} interference is activated.}
  \label{fig:scenario_cogloc}
\end{figure}

For the evaluation of this positioning approach, we use the seminar room scenario of the MeasureMINT database \cite{MeasureMINT2013}. The measurements allow for $5$ trajectories consisting of $1000$ agent positions with a \vu{1}{cm} spacing as shown in Fig.~\ref{fig:scenario_cogloc}. At each position, UWB measurements are available of the channel between the agent and the two anchors at the positions $\vm{a}_1^{(1)} = [0.5,7]\trans$ and $\vm{a}_1^{(2)} = [5.2,3.2]\trans$. The measurements have been performed using an M-sequence correlative channel sounder developed by \emph{Ilmsens}. This sounder provides measurements over approximately the FCC frequency range, from \vu{3-10}{GHz}. On anchor and agent sides, dipole-like  antennas made of Euro-cent coins have been used. They have an approximately uniform radiation pattern in azimuth plane and zeroes in the directions of floor and ceiling.

The chosen initial pulse duration is $\Tp=\vu{0.5}{ns}$ (corresponding to a bandwidth of $\vu{2}{GHz}$) and the center frequency is $\fc = \vu{7}{GHz}$. The {VA} for the anchors at the positions $\vm{a}_1^{(1)}$ and $\vm{a}_1^{(2)}$ were computed a-priori up to order $2$. The past window of agent positions for the SINR estimation is again chosen to be $w_\mathrm{past}=40$. For all simulations $30$ Monte Carlo runs were conducted. 

\subsection{Initial Experiment Setup}\label{ch7_firstexp}

For the sake of simplicity, we reduce the control parameters to just the carrier frequency $\vm{c}\timestep = f_{\mathrm{c},\timestepsym}$ for each {PAC} for all anchors and we fix the pulse duration $\Tp$. This means that the cognitive {MINT} system adaptively finds the carrier frequency $f_{\mathrm{c},\timestepsym}$ from {PAC} to {PAC} that yields the highest reward from the environment by maximizing the position-related information. Starting from the initial value $f_{\mathrm{c},1} = \vu{7}{GHz}$ (which represents the center of the measured bandwidth), the carrier frequency is adapted over time using the posterior entropy in \eqref{eq:entropstatetime}. 

The finite space of cognitive actions $\mathcal{C}$ contains the discrete frequency values bounded by the measured bandwidth, i.e. $f_{\mathrm{c},\timestepsym,i} \in \mathcal{C}$, where $i = 1,\dots,|\mathcal{C}|$. The frequency spacing between the frequency bins is equidistant, $\Delta \fc = f_{\mathrm{c},\timestepsym,i+1}-f_{\mathrm{c},\timestepsym,i}$. For the experiments, we haven chosen \vu{\Delta \fc = 50}{MHz}, considering the large signal bandwidth of \vu{2}{GHz}. The starting policy is defined as a uniform distribution $\pi_1(\vm{c}',\vm{c}) = \mathcal{U}(f_{\mathrm{c},\timestepsym,1}, f_{\mathrm{c},\timestepsym,|\mathcal{C}|})$ and the cost-to-go function is chosen to be $J_1(\vm{c}) = 0 \ \forall \vm{c}$. The size of the planning subspace is $|\mathcal{M}| = 20$; the size of $\mathcal{C}$ is $|\mathcal{C}| = 40$. 

\subsection{Discussion of Performance Results}\label{ch7_performance_discussion}

\subsubsection{Conventional {MINT}}\label{ch7_compconvMINT}

%------------------------------------CDFs---------------------------------
\begin{figure}[t!]
	\centering
	\psfrag{xaxis}{\small \textcolor{black}{\raisebox{-.3em}{\hspace{-1em}$\mathcal{P}(\pind{}) [m]$}}}
	\psfrag{yaxis}{\small \textcolor{black}{\raisebox{-.0em}{\hspace{0em}CDF}}}
	\includegraphics[width=1\columnwidth,keepaspectratio=true]{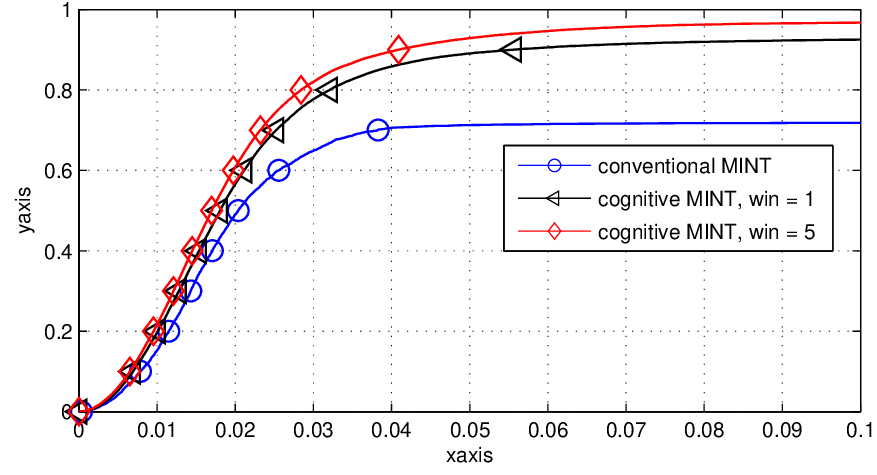}
	\caption{Performance {CDF} of the cognitive {MINT} algorithm using a smaller restricted set of {VA}. Visibilities of {VA} are computed using the {SINR} instead of optical ray-tracing.}
\label{fig:CDF_MINTUKFvsCogloc_lessinfo}
\end{figure}

Fig.~\ref{fig:CDF_MINTUKFvsCogloc_lessinfo} shows the overall position error {CDF} for ``conventional'' {MINT} (which assumes perfect floor plan knowledge) with and without cognitive waveform adaptation. To show the advantage of the cognitive {MINT} algorithm, a restricted set of {VA} is chosen and the visibilities of the {VA} are computed using the {SINR} instead of optical ray-tracing. As the {CDF} of ``conventional'' {MINT} indicates (blue line with circle marker), the tracking algorithm tends to diverge since too little position-related information is available. The black and the red lines show the overall position error {CDF} for cognitive {MINT} for a future horizon window of $l = 1$ and $l = 5$, respectively. As one can observe, the performance is significantly increased due to the cognitive waveform adaptation. This means that the cognitive {MINT} algorithm is able to increase the amount of position-related information by changing the sensing spectrum via the carrier frequency $f_{\mathrm{c},\timestepsym,i} \in \mathcal{A}$ to bands that carry more geometry-dependent information in the {MPC}. Another interesting observation of Fig.~\ref{fig:CDF_MINTUKFvsCogloc_lessinfo} is that an increase of the planning horizon results in an increased performance, confirming the correct functionality of the cognitive algorithm.

\subsubsection{Probabilistic {MINT}}\label{ch7_compprobMINT}

%------------------------------------CDFs---------------------------------
\begin{figure}[t!]
	\centering
	\psfrag{xaxis}{\small \textcolor{black}{\raisebox{-.3em}{\hspace{-1em}$\mathcal{P}(\pind{}) [m]$}}}
	\psfrag{yaxis}{\small \textcolor{black}{\raisebox{-.0em}{\hspace{0em}CDF}}}
	\includegraphics[width=1\columnwidth,keepaspectratio=true]{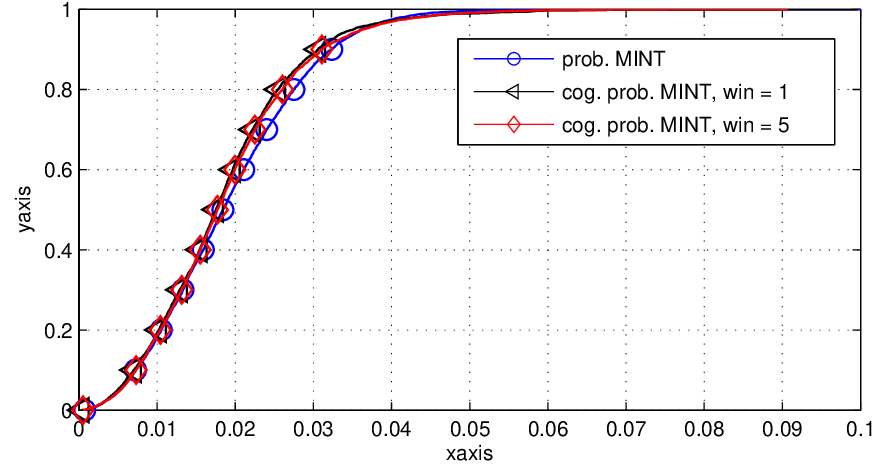}
	\caption{Performance {CDF} of cognitive probabilistic {MINT} using a smaller restricted set of {VA}. For probabilistic {MINT}, the visibilities of {VA} are always computed using the {SINR}.}
\label{fig:CDF_probMINTUKFvsCogloc_lessinfo}
\end{figure}

Fig.~\ref{fig:CDF_probMINTUKFvsCogloc_lessinfo} shows the overall position error {CDF} for probabilistic {MINT} with and without cognitive waveform adaptation. Uncertainties in the floor plan and wrong associations can be robustly handled due to the probabilistic treatment of {VA} and thus none of the individual trajectory runs diverges. The already achieved high accuracy and robustness of probabilistic {MINT} are the reasons that cognitive sensing leads to only a minor additional performance gain for this scenario. It is suspected that for lower bandwidth the performance gain induced by the cognitive probabilistic {MINT} should be much more distinct.     

\subsubsection{Probabilistic {MINT} with additional {DM} Interference}\label{ch7_compprobMINT_noise}

In the last setup, we additionally have added synthetic {DM} interference filtered at a carrier frequency \vu{f_c = 7}{GHz}, with a bandwidth of \vu{2}{GHz}. The {DM} parameters are chosen according to \cite{KaredalTWC2007} except for the {DM} power.                                                                                                                    The experiments were conducted with three levels of {DM} power, $\Omega_1 = 1.1615*10^{-9}$, $\Omega_1 = 5.8076*10^{-9}$ and $\Omega_1 = 1.1615*10^{-8}$.

%------------------------------------Signals---------------------------------
\begin{figure}[t!]
	\centering
	\psfragoffset{xaxis}{\footnotesize}{black}{-1}{-.3}{path delay $[m]$}
	\psfragoffset{A1}{\footnotesize}{black}{-1.2}{.2}{$|r\timestep^{(1)}(t)|$}
	\psfragoffset{A2}{\footnotesize}{black}{-1.2}{.2}{$|r\timestep^{(2)}(t)|$}
	\subfloat[Clean signals]{
		\includegraphics[width=1\columnwidth]{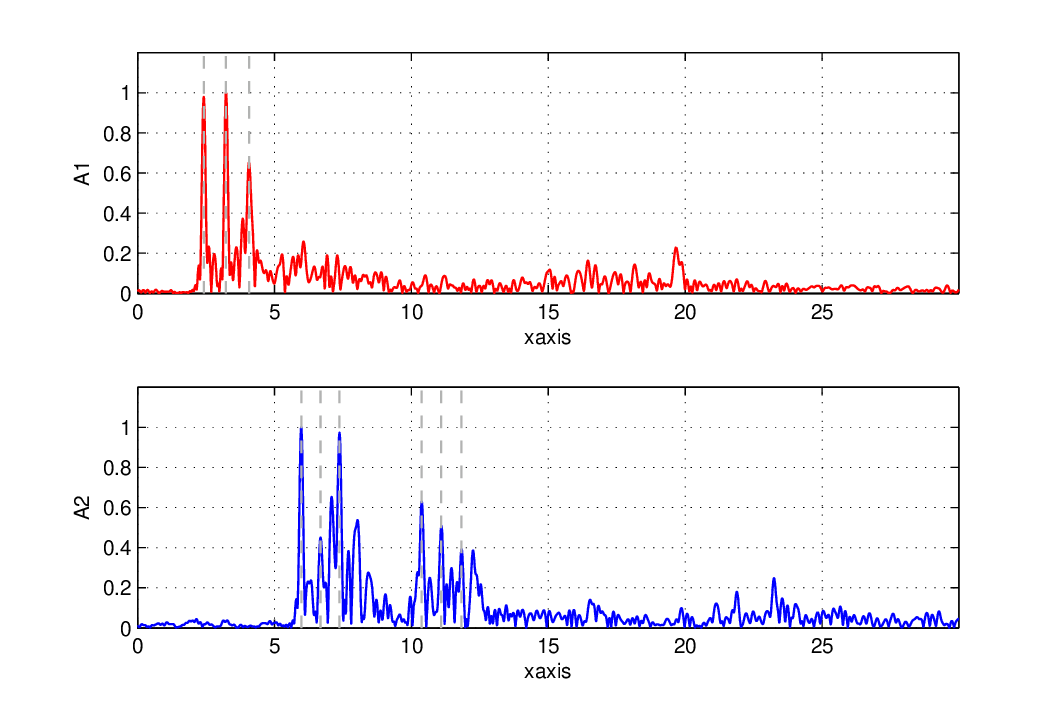}
		\label{fig:rx_signal_cogloc_clean} 
	}
	\hfill
	\subfloat[Noisy signal with {DM} power of $\Omega_1 = 1.1615*10^{-8}$]{
		\includegraphics[width=1\columnwidth]{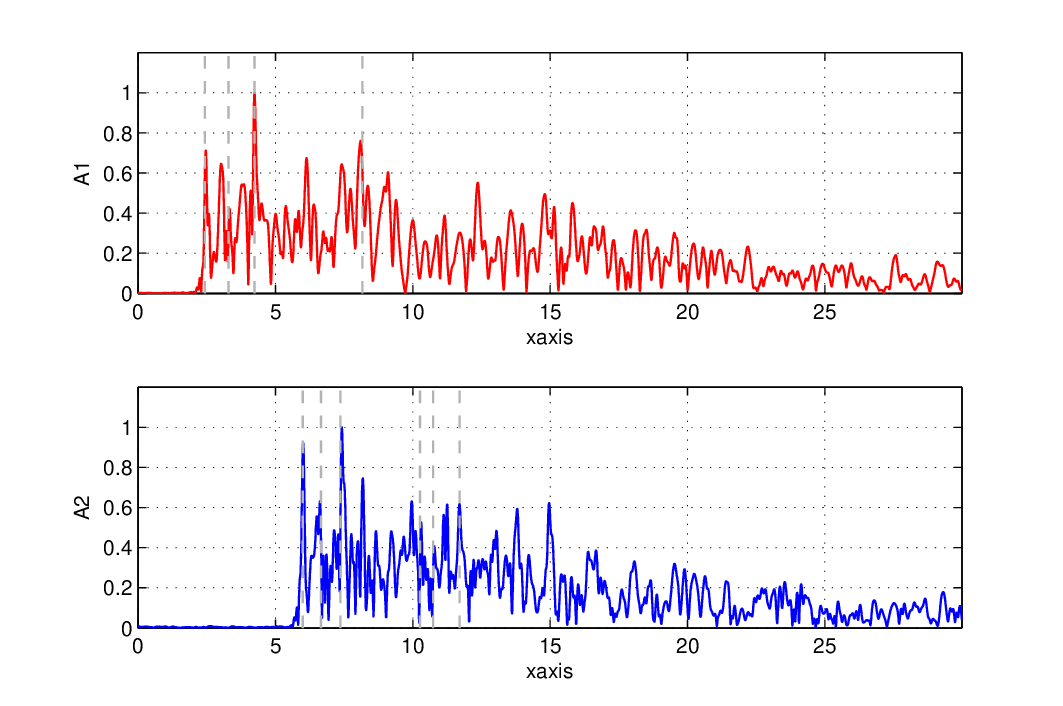}
		\label{fig:rx_signal_cogloc_disturb03}
	}
	\caption{Signals exchanged between agent and Anchors $1$ and $2$ for an example agent position. The gray lines represent the estimated delays of the {MPC}. Fig.~\ref{fig:rx_signal_cogloc_clean} shows the ``clean'' signal and Fig.~\ref{fig:rx_signal_cogloc_disturb03} the noisy signal.}
	\label{fig:rx_signal_cogloc}
\end{figure}

Fig.~\ref{fig:scenario_cogloc} illustrates the scenario used for the experiment. The black line represents the agent trajectory and the red part of it indicates the agent positions, where the {DM} interference is activated. Fig.~\ref{fig:rx_signal_cogloc} shows the signals exchanged between the agent and the Anchors $1$ and $2$ for one sample position. The ``clean'' signals are shown in Fig.~\ref{fig:rx_signal_cogloc_clean}, the noisy signal for {DM} power of $\Omega_1 = 1.1615*10^{-9}$ in Fig.~\ref{fig:rx_signal_cogloc_disturb03}. Looking at Fig.~\ref{fig:rx_signal_cogloc_disturb03} it is quite obvious that this level of {DM} represents already a severe interference. The justification of using such a interference noise model lies in the fact that it can describe many kinds of measurement modeling mismatches, e.g.\ if the anisotropy of the antenna pattern for different angle of arrivals is not considered. 

%%%%%%%%%%%%%%%%%%frequency over time%%%%%%%%%%%%%%%%%%%%%%%%%
\begin{figure}[t!]
	\psfrag{xaxis}{\small \textcolor{black}{\raisebox{-.3em}{\hspace{-2.5em} time index $\timestepsym$}}}
	\psfrag{yaxis}{\small \textcolor{black}{\raisebox{.4em}{\hspace{-2.5em} \vu{f}{[GHz]}}}}
	\centering
	\includegraphics[width=1\columnwidth]{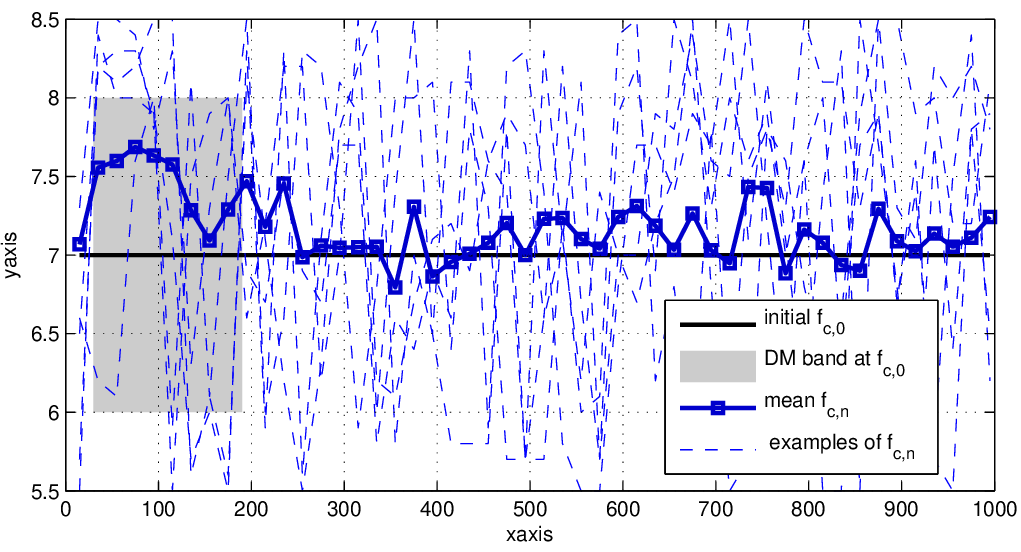}
 \caption{Mean carrier frequency for {DM} power $\Omega_1 = 1.1615*10^{-8}$. The black line denotes for the initial carrier frequency $f_{\mathrm{c},1}$ and the blue one the mean of the cognitively adapted carrier frequency $f_{\mathrm{c},\timestepsym}$. The blue dashed lines show a few example realizations of cognitively adapted carrier frequencies along different trajectories and for different Monte Carlo runs.}
\label{fig:f_opt_traj}
\end{figure}

Fig.~\ref{fig:f_opt_traj} illustrates the mean values of the cognitively adapted carrier frequency along one of the trajectories at {DM} power $\Omega_1 = 1.1615*10^{-8}$. The mean is computed using the $30$ Monte Carlo simulations of the experiment. The black line denotes the initial carrier frequency $f_{\mathrm{c},1}$ and the blue one the mean of the cognitively adapted carrier $f_{\mathrm{c},\timestepsym}$. The blue dashed lines show a few example realizations of cognitively adapted carrier frequencies along different trajectories and for different Monte Carlo runs. The figure shows quite clearly that the cognitive probabilistic {MINT} algorithm is avoiding (almost at all agent positions, where additional {DM} interference is present) carrier frequencies $f_{c,\timestepsym}$ near to the carrier of {DM}. 

%%%%%%%%%%%%%%%%%%entropy over time%%%%%%%%%%%%%%%%%%%%%%%%%
\begin{figure}[t!]
	\psfrag{xaxis}{\small \textcolor{black}{\raisebox{-.3em}{\hspace{-2.5em} time index $\timestepsym$}}}
	\psfrag{yaxis}{\small \textcolor{black}{\raisebox{.4em}{\hspace{-2.5em} Entropy}}}
	\centering
	\includegraphics[width=1\columnwidth]{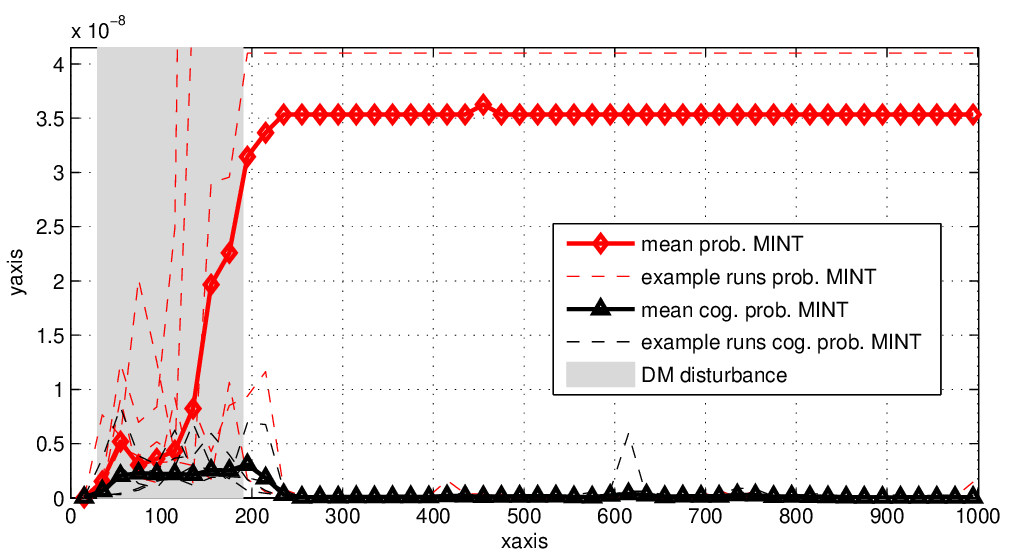}
 \caption{Mean entropy of probabilistic {MINT} and cognitive probabilistic {MINT} over time instances $\timestepsym$ for {DM} power $\Omega_1 = 1.1615*10^{-8}$. The red and black dashed lines show a few example entropy realizations along different trajectories and for different Monte Carlo runs.}
\label{fig:learning_curve}
\end{figure}

Fig.~\ref{fig:learning_curve} shows the according mean entropy values of probabilistic {MINT} (red line with diamond markers) and cognitive probabilistic {MINT} (black line with triangle markers) over time instances $\timestepsym$ for {DM} power $\Omega_1 = 1.1615*10^{-8}$. The red and black dashed lines show a few example entropy realizations along different trajectories and for different Monte Carlo runs. Before the noise disturbance starts the entropy of the probabilistic {MINT} algorithm is almost the same as of the cognitive probabilistic {MINT} algorithm. In the moment the disturbance is introduced, the entropy of the posterior increases. The cognitive probabilistic {MINT} algorithm then starts to change its carrier frequency $f_{c,\timestepsym}$ (as shown in Fig.~\ref{fig:f_opt_traj}) until the entropy is again reduced. This leads to an almost constant or even decreasing entropy even in the presence of a tremendous noise level (black line with triangle markers in Fig.~\ref{fig:learning_curve}). In contrast to that the probabilistic {MINT} algorithm without cognitive waveform adaptation starts to diverge after the disturbance is introduced and is not able to recover. This is indicated by the rapid increase of the entropy and stagnation at a large value shown in Fig.~\ref{fig:learning_curve} by the red line with diamond markers.   

%------------------------------------CDFs---------------------------------
\begin{figure}[!ht]
	\centering
	\psfrag{xaxis}{\small \textcolor{black}{\raisebox{-.3em}{\hspace{-1em}$\mathcal{P}(\pind{}) [m]$}}}
	\psfrag{yaxis}{\small \textcolor{black}{\raisebox{-.0em}{\hspace{0em}CDF}}}
	\includegraphics[width=1\columnwidth,keepaspectratio=true]{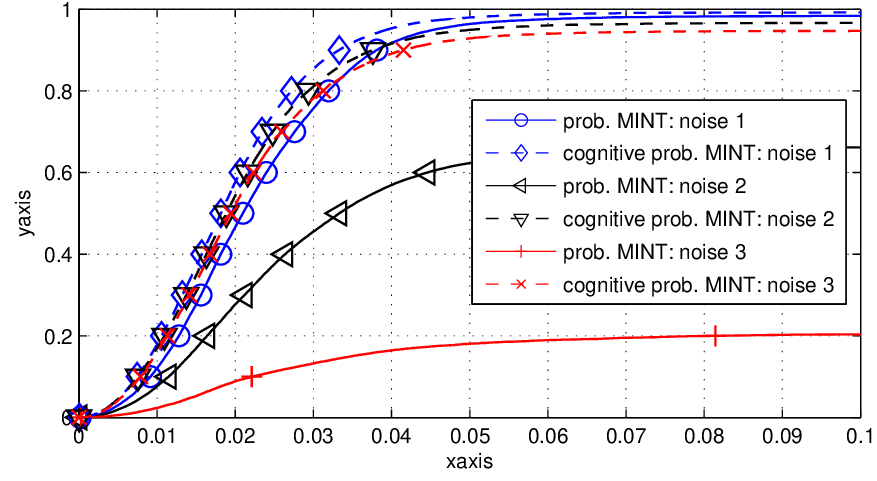}
	\caption{Performance {CDF} of the cognitive probabilistic {MINT} algorithm with introducing a disturbance at three different noise levels along a certain part of the trajectory. Noise $1$ corresponds to {DM} with $\Omega_1 = 1.1615*10^{-9}$, Noise $2$ with power $\Omega_1 = 5.8076*10^{-9}$ and with power $\Omega_1 = 1.1615*10^{-8}$}
\label{fig:probMINTUKFvsCogloc_lessinfo_noise}
\end{figure}

This result is confirmed by looking at the performance {CDF} of the agent position error shown in Fig.~\ref{fig:probMINTUKFvsCogloc_lessinfo_noise}. This comparison between probabilistic {MINT} and cognitive probabilistic {MINT} illustrates the powerful property of the cognitive algorithm to separate relevant from irrelevant information using adaptation of the control parameter $f_{c,\timestepsym}$ to avoid the noisy frequency band of the signal. The probabilistic {MINT} algorithm without waveform adaptation tends to diverge under such harsh conditions as depicted by {CDF} drawn with solid lines. In contrast to this, the cognitive {MINT} algorithm overcomes these impairments, leading again to a \emph{robust} behavior as depicted by {CDF} drawn with dashed lines.

\balance

\IEEEtriggeratref{0}
\bibliographystyle{IEEEtran}
% \bibliography{IEEEabrv,Bib_Jabref}
\bibliography{IEEEabrv,Bib_PhD}

\begin{thebibliography}{10}
\providecommand{\url}[1]{#1}
\csname url@samestyle\endcsname
\providecommand{\newblock}{\relax}
\providecommand{\bibinfo}[2]{#2}
\providecommand{\BIBentrySTDinterwordspacing}{\spaceskip=0pt\relax}
\providecommand{\BIBentryALTinterwordstretchfactor}{4}
\providecommand{\BIBentryALTinterwordspacing}{\spaceskip=\fontdimen2\font plus
\BIBentryALTinterwordstretchfactor\fontdimen3\font minus
  \fontdimen4\font\relax}
\providecommand{\BIBforeignlanguage}[2]{{%
\expandafter\ifx\csname l@#1\endcsname\relax
\typeout{** WARNING: IEEEtran.bst: No hyphenation pattern has been}%
\typeout{** loaded for the language `#1'. Using the pattern for}%
\typeout{** the default language instead.}%
\else
\language=\csname l@#1\endcsname
\fi
#2}}
\providecommand{\BIBdecl}{\relax}
\BIBdecl

\bibitem{Fuster2009}
J.~M. Fuster, \emph{Cortex and Mind - Unifying Cognition}.\hskip 1em plus 0.5em
  minus 0.4em\relax Oxford University Press, 2003.

\bibitem{MeissnerPhD2014}
P.~Meissner, ``Multipath-{Assisted Indoor Positioning},'' Ph.D. dissertation,
  Graz University of Technology, 2014.

\bibitem{LeitingerJSAC2015}
E.~Leitinger, P.~Meissner, C.~Rudisser, G.~Dumphart, and K.~Witrisal,
  ``{Evaluation of Position-Related Information in Multipath Components for
  Indoor Positioning},'' \emph{IEEE Journal on Selected Areas in
  Communications}, vol.~33, no.~11, pp. 2313--2328, Nov 2015.

\bibitem{LeitingerPhD2016}
E.~Leitinger, ``{Cognitive Indoor Positioning and Tracking using Multipath
  Channel Information},'' Ph.D. dissertation, Graz University of Technology,
  2016.

\bibitem{LeitingerGNSS2016}
E.~Leitinger, M.~F., P.~Meissner, K.~Witrisal, and F.~Hlawatsch, ``{Belief
  Propagation based Joint Probabilistic Data Association for Multipath-Assisted
  Indoor Navigation and Tracking},'' in \emph{2016 International Conference on
  Localization and GNSS (ICL-GNSS)}, June 2016.

\bibitem{HaykinPROC2012a}
S.~Haykin, Y.~Xue, and P.~Setoodeh, ``{Cognitive Radar: Step Toward Bridging
  the Gap Between Neuroscience and Engineering},'' \emph{Proceedings of the
  IEEE}, vol. 100, no.~11, pp. 3102 --3130, nov. 2012.

\bibitem{HaykinPROC2012b}
S.~Haykin, M.~Fatemi, P.~Setoodeh, and Y.~Xue, ``{Cognitive Control},''
  \emph{Proceedings of the IEEE}, vol. 100, no.~12, pp. 3156 --3169, dec. 2012.

\bibitem{Fatemi2014}
M.~Fatemi and S.~Haykin, ``{Cognitive Control: Theory and Application},''
  \emph{Access, IEEE}, vol.~2, pp. 698--710, 2014.

\bibitem{AmiriNC2014}
A.~Amiri and S.~Haykin, ``{Improved Sparse Coding Under the Influence of
  Perceptual Attention},'' \emph{Neural Comput.}, vol.~26, no.~2, pp. 377--420,
  Feb. 2014.

\bibitem{HaykinPROC2014}
S.~Haykin and J.~Fuster, ``{On Cognitive Dynamic Systems: Cognitive
  Neuroscience and Engineering Learning From Each Other},'' \emph{Proceedings
  of the IEEE}, vol. 102, no.~4, pp. 608--628, April 2014.

\bibitem{Pearl1988}
J.~Pearl, \emph{Probabilistic Reasoning in Intelligent Systems: Networks of
  Plausible Inference}.\hskip 1em plus 0.5em minus 0.4em\relax San Francisco,
  CA, USA: Morgan Kaufmann Publishers Inc., 1988.

\bibitem{Gregory2005_BLD}
P.~Gregory, \emph{Bayesian Logical Data Analysis for the Physical
  Sciences}.\hskip 1em plus 0.5em minus 0.4em\relax New York, NY, USA:
  Cambridge University Press, 2005.

\bibitem{SiviaSkilling2006}
D.~S. Sivia and J.~Skilling, \emph{Data analysis : a Bayesian tutorial}, ser.
  Oxford science publications.\hskip 1em plus 0.5em minus 0.4em\relax Oxford,
  New York: Oxford University Press, 2006.

\bibitem{Borish1984}
J.~Borish, ``{Extension of the Image Model to arbitrary Polyhedra},'' \emph{The
  Journal of the Acoustical Society of America}, March 1984.

\bibitem{KunischICUWB2003}
J.~Kunisch and J.~Pamp, ``{An Ultra-Wideband space-variant Multipath Indoor
  Radio Channel Model},'' in \emph{Ultra Wideband Systems and Technologies,
  2003 IEEE Conference on}, Nov 2003, pp. 290--294.

\bibitem{LeitingerICC2015}
E.~Leitinger, P.~Meissner, M.~Lafer, and K.~Witrisal, ``{Simultaneous
  Localization and Mapping using Multipath Channel Information},'' in
  \emph{2015 IEEE International Conference on Communications Workshops (ICC)},
  London, UK, June 2015, pp. 754--760.

\bibitem{LeitingerICC2017}
E.~Leitinger, F.~Meyer, F.~Tufvesson, and K.~Witrisal, ``Factor graph based
  simultaneous localization and mapping using multipath channel information,''
  in \emph{Proc. IEEE ICCW-17}, Paris, France, May 2017, pp. 652--658.

\bibitem{LeitingerTWC2019}
E.~{Leitinger}, F.~{Meyer}, F.~{Hlawatsch}, K.~{Witrisal}, F.~{Tufvesson}, and
  M.~Z. {Win}, ``A belief propagation algorithm for multipath-based {SLAM},''
  \emph{{IEEE} Trans. Wireless Commun.}, vol.~18, no.~12, pp. 5613--5629, Dec.
  2019.

\bibitem{LeitingerICC2019}
E.~{Leitinger}, S.~{Grebien}, and K.~{Witrisal}, ``Multipath-based {SLAM}
  exploiting {AoA} and amplitude information,'' in \emph{Proc. IEEE ICCW-19},
  Shanghai, China, May 2019, pp. 1--7.

\bibitem{Kershaw1994}
D.~Kershaw and R.~Evans, ``Optimal waveform selection for tracking systems,''
  \emph{Information Theory, IEEE Transactions on}, vol.~40, no.~5, pp. 1536
  --1550, sep 1994.

\bibitem{Haykin2011}
S.~Haykin, A.~Zia, Y.~Xue, and I.~Arasaratnam, ``{Control Theoretic Approach to
  Tracking Radar: First step towards cognition},'' \emph{Digital Signal
  Processing}, vol.~21, no.~5, pp. 576 -- 585, 2011.

\bibitem{BellJSTSP2015}
K.~Bell, C.~Baker, G.~Smith, J.~Johnson, and M.~Rangaswamy, ``Cognitive radar
  framework for target detection and tracking,'' \emph{Selected Topics in
  Signal Processing, IEEE Journal of}, vol.~9, no.~8, pp. 1427--1439, Dec 2015.

\bibitem{HoffmannTAC2010}
G.~Hoffmann and C.~Tomlin, ``{Mobile Sensor Network Control Using Mutual
  Information Methods and Particle Filters},'' \emph{Automatic Control, IEEE
  Transactions on}, vol.~55, no.~1, pp. 32--47, Jan 2010.

\bibitem{JulianJRR2012}
\BIBentryALTinterwordspacing
B.~J. Julian, M.~Angermann, M.~Schwager, and D.~Rus, ``{Distributed Robotic
  Sensor Networks: An Information-theoretic Approach},'' \emph{Int. J. Rob.
  Res.}, vol.~31, no.~10, pp. 1134--1154, Sep. 2012. [Online]. Available:
  \url{http://dx.doi.org/10.1177/0278364912452675}
\BIBentrySTDinterwordspacing

\bibitem{Chaloner1995}
\BIBentryALTinterwordspacing
K.~Chaloner and I.~Verdinelli, ``\BIBforeignlanguage{English}{{Bayesian
  Experimental Design: A Review}},''
  \emph{\BIBforeignlanguage{English}{Statistical Science}}, vol.~10, no.~3, pp.
  pp. 273--304, 1995. [Online]. Available:
  \url{http://www.jstor.org/stable/2246015}
\BIBentrySTDinterwordspacing

\bibitem{Cover2006}
T.~M. Cover and J.~A. Thomas, \emph{Elements of Information Theory (Wiley
  Series in Telecommunications and Signal Processing)}.\hskip 1em plus 0.5em
  minus 0.4em\relax Wiley-Interscience, 2006.

\bibitem{VanTrees1968}
H.~L. Van~Trees, \emph{Detection, Estimation and Modulation, Part {I}}.\hskip
  1em plus 0.5em minus 0.4em\relax Wiley Press, 1968.

\bibitem{Kay1993}
S.~Kay, \emph{Fundamentals of Statistical Signal Processing: Estimation
  Theory}.\hskip 1em plus 0.5em minus 0.4em\relax Prentice Hall Signal
  Processing Series, 1993.

\bibitem{MeissnerWCL2014}
P.~Meissner, E.~Leitinger, and K.~Witrisal, ``{UWB} for robust indoor tracking:
  Weighting of multipath components for efficient estimation,'' \emph{Wireless
  Communications Letters, IEEE}, vol.~3, no.~5, pp. 501--504, Oct 2014.

\bibitem{MichelusiTSP2012part1}
N.~Michelusi, U.~Mitra, A.~Molisch, and M.~Zorzi, ``{UWB Sparse/Diffuse
  Channels, {Part I}: Channel Models and Bayesian Estimators},'' \emph{Signal
  Processing, IEEE Transactions on}, vol.~60, no.~10, pp. 5307--5319, 2012.

\bibitem{MolischTPROC2009}
A.~Molisch, ``{Ultra-Wide-Band Propagation Channels},'' \emph{Proceedings of
  the IEEE}, vol.~97, no.~2, pp. 353--371, Feb. 2009.

\bibitem{GrebienLeitingerTSP2021}
S.~Grebien, E.~Leitinger, K.~Witrisal, and B.~H. Fleury, ``Super-resolution
  channel estimation including the dense multipath component --- {A} sparse
  variational {Bayesian} approach,'' 2021, in preperation.

\bibitem{WitrisalSPM2016}
K.~Witrisal, P.~Meissner, E.~Leitinger, Y.~Shen, C.~Gustafson, F.~Tufvesson,
  K.~Haneda, D.~Dardari, A.~F. Molisch, A.~Conti, and M.~Z. Win,
  ``{High-Accuracy Localization for Assisted Living: 5G systems will turn
  multipath channels from foe to friend},'' \emph{IEEE Signal Processing
  Magazine}, vol.~33, no.~2, pp. 59--70, March 2016.

\bibitem{ShenTIT2010part1}
Y.~Shen and M.~Win, ``{Fundamental Limits of Wideband Localization; Part {I}: A
  General Framework},'' \emph{Information Theory, IEEE Transactions on},
  vol.~56, no.~10, pp. 4956--4980, Oct. 2010.

\bibitem{ShenTIT2010part2}
Y.~Shen, H.~Wymeersch, and M.~Win, ``{Fundamental Limits of Wideband
  Localization; Part {II}: Cooperative Networks},'' \emph{Information Theory,
  IEEE Transactions on}, vol.~56, no.~10, pp. 4981 --5000, Oct. 2010.

\bibitem{MeyerProc2018}
F.~Meyer, T.~Kropfreiter, J.~L. Williams, R.~Lau, F.~Hlawatsch, P.~Braca, and
  M.~Z. Win, ``Message passing algorithms for scalable multitarget tracking,''
  \emph{Proc. {IEEE}}, vol. 106, no.~2, pp. 221--259, Feb. 2018.

\bibitem{MeyerIFC2015}
F.~Meyer, P.~Braca, P.~Willett, and F.~Hlawatsch, ``{Scalable Multitarget
  Tracking using Multiple Sensors: A belief propagation approach},'' in
  \emph{Information Fusion (Fusion), 2015 18th International Conference on},
  July 2015, pp. 1778--1785.

\bibitem{barShalom95}
Y.~Bar-Shalom and X.-R. Li, \emph{{Multitarget-Multisensor Tracking :
  Principles and Techniques}}.\hskip 1em plus 0.5em minus 0.4em\relax Storrs,
  CT: Yaakov Bar-Shalom, 1995.

\bibitem{vermaak05}
J.~Vermaak, S.~J. Godsill, and P.~Perez, ``{Monte Carlo filtering for multi
  target tracking and data association},'' vol.~41, no.~1, pp. 309--332, Jan.
  2005.

\bibitem{kay1998}
S.~Kay, \emph{Fundamentals of Statistical Signal Processing: Detection
  Theory}.\hskip 1em plus 0.5em minus 0.4em\relax Prentice Hall Signal
  Processing Series, 1998.

\bibitem{WymeerschICC2013}
H.~Wymeersch, ``{The Impact of Cooperative Localization on Achieving
  higher-level Goals},'' in \emph{Communications Workshops (ICC), 2013 IEEE
  International Conference on}, June 2013, pp. 1--5.

\bibitem{MeyerJSAC2015}
F.~Meyer, H.~Wymeersch, M.~Frohle, and F.~Hlawatsch, ``{Distributed Estimation
  With Information-Seeking Control in Agent Networks},'' \emph{Selected Areas
  in Communications, IEEE Journal on}, vol.~33, no.~11, pp. 2439--2456, Nov
  2015.

\bibitem{GrocholskyPhD2002}
B.~Grocholsky and B.~Grocholsky, ``{Information-Theoretic Control of Multiple
  Sensor Platforms},'' Ph.D. dissertation, Department of Aerospace, Mechatronic
  and Mechanical Engineering, 2002.

\bibitem{Shannon1948}
C.~Shannon, ``{A Mathematical Theory of Communication},'' \emph{Bell System
  Technical Journal, The}, vol.~27, no.~4, pp. 623--656, Oct 1948.

\bibitem{HaykinBook2012}
S.~Haykin, \emph{Cognitive Dynamic Systems: Perception-action Cycle, Radar and
  Radio}.\hskip 1em plus 0.5em minus 0.4em\relax New York, NY, USA: Cambridge
  University Press, 2012.

\bibitem{Bellman1957}
R.~Bellman, \emph{Dynamic Programming}.\hskip 1em plus 0.5em minus 0.4em\relax
  Princeton, NJ, USA: Princeton University Press, 1957.

\bibitem{Bertsekas2000}
D.~P. Bertsekas, \emph{Dynamic Programming and Optimal Control}.\hskip 1em plus
  0.5em minus 0.4em\relax Athena Scientific, 2000.

\bibitem{Sutton1998}
R.~S. Sutton and A.~G. Barto, \emph{Introduction to Reinforcement Learning},
  1st~ed.\hskip 1em plus 0.5em minus 0.4em\relax Cambridge, MA, USA: MIT Press,
  1998.

\bibitem{Lazaric07reinforcementlearning}
A.~Lazaric, M.~Restelli, and A.~Bonarini, ``Reinforcement learning in
  continuous action spaces through sequential monte carlo methods,'' in
  \emph{Advances in Neural Information Processing Systems}, 2007.

\bibitem{Sahinoglu2008}
Z.~Sahinoglu, S.~Gezici, and I.~Guvenc, \emph{Ultra-wideband {Positioning
  Systems -- Theoretical Limits, Ranging Algorithms and Protocols}}.\hskip 1em
  plus 0.5em minus 0.4em\relax Cambridge University Press, 2008.

\bibitem{MeasureMINT2013}
\BIBentryALTinterwordspacing
P.~Meissner, E.~Leitinger, M.~Lafer, and K.~Witrisal, ``M{easureMINT UWB
  database},'' www.spsc.tugraz.at/tools/UWBmeasurements, 2013, {Publicly
  available database of UWB indoor channel measurements}. [Online]. Available:
  \url{www.spsc.tugraz.at/tools/UWBmeasurements}
\BIBentrySTDinterwordspacing

\bibitem{KaredalTWC2007}
J.~Karedal, S.~Wyne, P.~Almers, F.~Tufvesson, and A.~Molisch, ``{A
  Measurement-Based Statistical Model for Industrial Ultra-Wideband
  Channels},'' \emph{Wireless Communications, IEEE Transactions on}, vol.~6,
  no.~8, pp. 3028--3037, Aug. 2007.

\end{thebibliography}

\end{document}